Self-move and Other-move:

Quantum Categorical Foundations of Japanese

Ryder Dale Walton

Capitol Technical University

**Abstract**

The purpose of this work is to contribute toward the larger goal of creating a Quantum Natural Language Processing (QNLP) translator program. This work contributes original diagrammatic representations of the Japanese language based on prior work that accomplished on the English language based on category theory. The germane differences between the English and Japanese languages are emphasized to help address English language bias in the current body of research. Additionally, topological principles of these diagrams and many potential avenues for further research are proposed. Why is this endeavor important? Hundreds of languages have developed over the course of millennia coinciding with the evolution of human interaction across time and geographic location. These languages are foundational to human survival, experience, flourishing, and living the good life. They are also, however, the strongest barrier between people groups. Over the last several decades, advancements in Natural Language Processing (NLP) have made it easier to bridge the gap between individuals who do not share a common language or culture. Tools like Google Translate and DeepL make it easier than ever before to share our experiences with people globally. Nevertheless, these tools are still inadequate as they fail to convey our ideas across the language barrier fluently, leaving people feeling anxious and embarrassed. This is particularly true of languages born out of substantially different cultures, such as English and Japanese. Quantum computers offer the best chance to achieve translation fluency in that they are better suited to simulating the natural world and natural phenomenon such as natural speech.

*Keywords:* category theory, DisCoCat, DisCoCirc, Japanese grammar, English grammar, translation, topology, Quantum Natural Language Processing, Natural Language Processing

## Why Quantum Computers?

*Natural Language Processing* (NLP) is a field of much interest in computer science and continues to produce significant new technologies. Indeed, the ongoing success of NLP cannot be oversold. Voice assistants—Siri and Google Assistant—and smart home hubs—Amazon Alexa, Google Home, and Apple HomePod—rely on NLP for all their command inputs and basic functions. This includes reception of audio, generating text, parsing said text into tokens, processing the tokens, finding search results, and performing other various commands. If classical computers provide all these NLP benefits and successes, why spend the time to research what quantum computers can do?

**The Problem**

Science Fiction authors and fans have long mused over the concept of a universal language translator. In fact, even the significantly less ambitious endeavor of producing a fluent translator from one source language to another target language seems like magic. Why? Frankly, NLP, despite its resounding success and significant forward momentum, continues to fail to produce fluent translations. Consider Google Translate or DeepL. These machine translators provide a translation that is good enough to allow people from different cultures to communicate with one another. That alone is a modern marvel, but if you ask anyone who has used such a translator if the translations sounded natural—if they were fluent—at all, then they would simply respond with a "No". Again, why? It turns out that the translators do not know anything about culture. Further, they do not know anything about extra-textual context. Even worse, they do not know anything about grammar. Thus, ultimately, they do not know anything about language. Machine translators do know an awful lot about corpora, probability, and "bags of words" though, which is enough to be very effective and reasonably accurate with translations of single sentences (Coecke et al., 2020). However, it is not enough to be fluent. It is, therefore, still necessary to rely on human translators in most situations where trans-cultural communication is required. Machine translators are simply not capable enough to replace human translators in corporate meetings, travel tours, or long interpersonal interactions. The present study aims to provide one step toward solving this problem by building on research performed on English grammar in the field of *Quantum Natural Language Processing* (QNLP), extending that research to Japanese grammar.

But why would quantizing NLP provide a potential solution to the fluency problem? Well, language is a natural phenomenon that has developed organically in human brains and in human cultures over millennia. Since language is a natural phenomenon, creating approaches to simulate grammar, like simulating quantum-sized problems—molecular systems and chemical compounds—, is one potential path forward to improve machine language processing capabilities.[1] In fact, the underlying mathematics of both reduce to the same core category theoretic representation, namely compact closed categories (Zeng & Coecke, 2016). The main problem associated with modeling languages, both grammar and semantics, is that it is far too resource intensive to be feasible on classical computers. This, then, is precisely why quantum computers fit the bill. Modeling processes require many, many variables and have many, many possible outcomes. Such problems are said to be "quantum-native" because quantum computers provide a pathway to unlocking the needed resources that classical computers can never attain (QNLP, 2019). Namely, the tensor product of many qubits provides an exponential space advantage over classical bits (Meichanetzidis et al.,

---

[1] Scott Aaronson notes that simulation of quantum systems is one of the more obvious uses of quantum computation (2013).

2021). Suddenly, with quantum hardware, machine-generated translation fluency enters the realm of the possible because modeling of grammar and semantics becomes possible (Zeng & Coecke, 2016). Other benefits include potential Grover's algorithm-style quadratic speedups in processing (QNLP, 2019); and that quantum circuits are natively vectors or matrices (Coecke et al., 2018a). Particularly, working with density matrices allows for efficient preservation of more data, which means the system can track and manipulate more state data (Coecke & Meichanetzidis, 2020). These stated benefits are only a few of the known benefits of using quantum computers for language processing. The possibility for more benefits certainly exists and is a topic of continuing research.

**Figure 0**

*Comparison of a General NLP Pipeline with a General QNLP Pipeline*

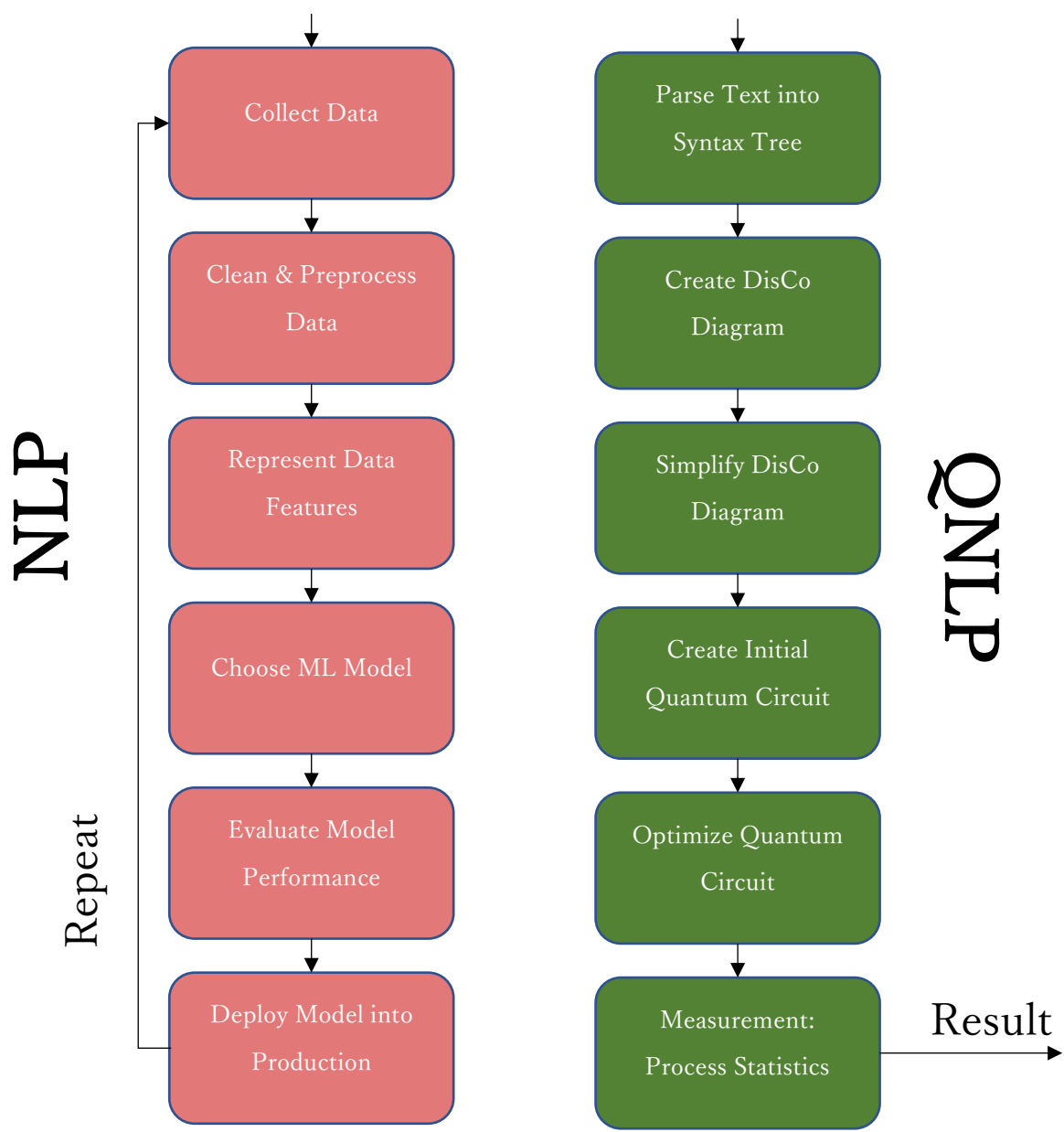

*Note.* This figure is original but based on work from a few different sources (Chen, 2020; Lorenz et al., 2021; Meichanetzidis et al., 2021). Because the NLP pipeline relies on Machine Learning (ML), there is a considerable amount of trial and error in each step. Different representations of data, machine learning models, evaluation methods,

preprocessing techniques, and the like all impact the performance of the model. Additionally, as new information is made available, that information must be fed back into the model to retrain it. The QNLP approach here, on the other hand, can leverage grammar, greatly simplifying the number of choices. Unfortunately, creating a quantum circuit from a DisCoCat diagram is not a solved problem and relies on ansatz and optimization. Still, having a set grammar eliminates the need for ML to be trained over a bag of words or any other approach. This is a great boon.

    The purpose of the present work, then, is to provide a necessary step forward for translating English into Japanese with a high degree of *accuracy*—whether the intended meaning was conveyed correctly—and *fluency*—whether the meaning was conveyed naturally. The present study differs from prior research in that it not only aims to contribute to QNLP research, but also intends to cover blind spots in the literature that may exist because of English language bias.[2] To achieve this, diagrammatic reasoning, which is based on category theory, will be used throughout the work to contribute new

---

[2] Abbaszade et al. assert that studying other languages is an important area of further research in QNLP (2021).

diagrams to the field to aid further research while specifying general topological principles for said diagrams.

## Background: Category Theory

All *categories* are made up of two types of entities, "*objects* and *morphisms*" (van de Wetering, 2020). Objects are the subjects of the category theory and morphisms are the legal operations on said subjects. An example is particularly enlightening because, to many, category theory may be unfamiliar territory. However, all current and former students of mathematics are familiar with this example, most basic category: Sets. In sets, objects are the sets themselves, and the morphisms are functions that map one set to another.

Categories require two basic operations. First, categories must have *sequential composition*, which must be associative in nature. Suppose the following morphisms exist in a basic category: $f: A \rightarrow B$ and $g: B \rightarrow C$. In relation to sets, these represent a transformation from the set on the left to the set on the right. Since sequential composition must be upheld to be a category, it must be possible to nest these two morphisms and create a new one. This nesting produces the following result: $g \circ f: A \rightarrow C$. This can be read as "$g$ after $f$" as a sequential composition represents morphemes that occur on objects one after the other (Coecke & Kissinger, 2017).

Associativity of sequential compositions means that the following must also be true: $h \circ (g \circ f) = (h \circ g) \circ f$. In other words, regardless of the order in which the morphemes are calculated produces the same result.

Secondly, categories must have an *identity* morphism. The identity morphism is quite simple: $id_A : A \rightarrow A$. Simply put, the identity morphism maps the object back onto itself. In the case of sets, the function would produce no change. When the identity morphism sequentially composes with another morphism, the result is as follows: $id_A \circ f = f$. Thus, sequentially composing any morphism with the identity morphism results in no change to the object. Nevertheless, the identity must hold for category theory broadly to hold.

While the above features of category theory are sufficient for working with sets, they are inadequate for working with both languages and quantum mechanics. The first missing piece is a way to represent actions happening simultaneously. Recall that sequential composition represents only sequential action—things happening one after the other. There must also be some way to represents actions happening at the same time. Fortunately, this is precisely what "monoidal" categories provide (van de Wetering, 2020). Monoidal categories do this by using the tensor product—a special quantum mechanical operator that denotes an addition of two subsystems, like vector spaces—to represent the sum of the two objects: $A \otimes B := A + B$. This expression provides *parallel composition* because there is no change to object A or change to object B. It simply says that within the expanse of the object both A and B are represented. Now, is parallel composition associative? Is $A \otimes B = B \otimes A$? This is not necessarily the case in a monoidal category, but it is true for a symmetric monoidal category (van de Wetering, 2020). The category is said to be symmetric because $A \otimes B$ and $B \otimes A$ are necessarily isomorphic. In this case, then, the order of the objects does not matter when they are composed in parallel.

The second missing piece is less intuitive. For a category theory to be sufficient for quantum mechanics, there must also be a way to transform inputs to outputs and outputs to inputs. When such a structure is present and the category is symmetric, the category is called "compact closed" (van de Wetering, 2020). Why is this necessary? For working with quantum mechanical ideas, the idea of non-separability must be addressed. Essentially, this means that there must be some graphical component that represents quantum entanglement because quantum theory is founded on the idea that somethings cannot be divided or separated—contrary to the assumptions of classical physics. To accomplish this goal, it turns out that the graph must allow inputs to be connected to other inputs and outputs connected to other outputs (Coecke & Kissinger, 2017). This concept is easier to depict graphically and will be discussed further in the next section. Suffice it to say for now that a compact closed symmetric monoidal category, or compact closed category for short, is sufficient for the present study.[3] It is, after all, "the fundamental structure in categorical quantum mechanics" (van de Wetering, 2020).

But what does this have to do with language? It turns out the compact closed categories that underlie quantum mechanics also underlie algebraic linguistics (Coecke, 2020). Since both quantum mechanics and algebraic linguistics boil down to compact closed categories, this reinforces the earlier assertion that language problems are quantum native. More discussion follows, but for now know that words correspond to quantum states, and grammatical structures correspond to quantum measurements (Coecke et al, 2021). Given this revelation, one can approach quantum problems and linguistic problems from the same perspective using the same tools. In fact, using these tools makes the preprocessing of data into a quantum friendly structure quite easy (O'Riordan et a., 2020).

---

[3] Compact closed categories are a special case of rigid categories (Meichanetzidis et al., 2021). Additionally, Zeng and Coecke use dagger compact categories as the basis of their work, which include the involutive functor, which is drawn as a dagger. This functor performs the identity operation on objects, but morphisms are mapped to their adjoints (2016). There is much more that could be discussed in this regard from the perspective of category theory, which is a branch of mathematics. For the current discussion, this level of understanding suffices.

**The Software Tool**

When it comes to category theory, one key software tool is DisCoPy, which is a portmanteau of "*Cat*egorical *Co*mpositional *Dis*tributional" (*DisCoCat*[4]) and Python (Coecke, 2020; de Felice et al., 2021). Before continuing, it is worth mentioning that DisCoCat and *Cir*cuit-shaped *Co*mpositional *Dis*tributional (*DisCoCirc*) are two related diagrammatic representations of category theoretical concepts that follow in this section (Coecke, 2020). Both kinds of diagrams can be referred to simultaneously as Compositional Distributional (*DisCo*) diagrams.[5] DisCoCat is used when determining whether a single sentence is grammatical and determining the meaning of a single sentence. DisCoCirc is used across a larger body of a text, in which each sentence is a process with inputs and outputs that are nouns, to track the updating of said nouns' meanings in said text (Coecke, 2020). Now, DisCoPy is a Python package that allows for easy creation of *string diagrams*, which are diagrammatic representations of morphisms in compact closed categories. This essentially means that DisCoPy allows for the creation of diagrams that are potentially useful for studying natural languages. Now, string diagrams are particularly helpful when working with category theories because of the over-abundance of symbols and expressions required to describe categories. The vernacular of category theory can also be quite overwhelming for the uninitiated. However, diagrammatic representations alleviate all this pain.

---

[4] The order of the terms in the DisCoCat is in reverse because the acronym follows the same logic as sequential composition. Stated plainly, it is "Distributional after Compositional after Categorical".

[5] The idea of distributional structure is decades old and is based on the idea that grammar enforces order, restrictions, and relative occurrences of elements (Harris, 1954). This suggests that language can be represented by distributionally and is a cornerstone of DisCo diagrams.

For a first example, consider the morphemes *f* and *g* mentioned previously. The former was simply defined as some operation that changes object A to object B: $f: A \to B$. Diagrammatically, DisCoPy depicts this as a box—morpheme—with two wires—objects. Figure 1 is an example.

**Figure 1**

*f Depicted Diagrammatically Using DisCoPy*

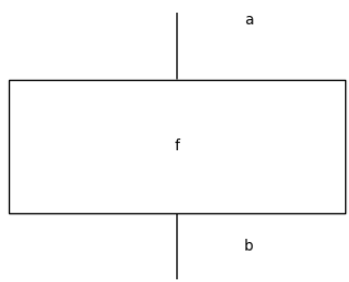

Throughout this work, the standard will be that the input, or domain, is represented at the top of the diagram. The bottom of the diagram is the output, or codomain. Thus, this diagram represents some input A that undergoes some transformation process into the output B. It could not be any simpler. $g: B \rightarrow C$ defines a morpheme that changes B into C and is depicted similarly in Figure 2. While this may seem trivial at first, there is one difference between *f* and *g* that is worth noting. The *base types* do not match. A, B, and C all represent something different. Thus, *f* and *g* may not be equivalent processes.

**Figure 2**

*g Depicted Diagrammatically Using DisCoPy*

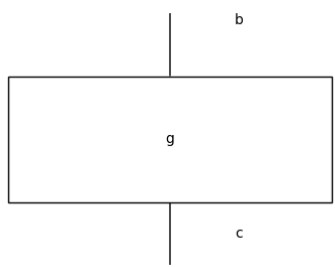

Furthermore, these processes may be composed together. Recall $g \circ f: A \to C$, which denotes the process $g$ taking place after the process $f$ concludes. Diagrammatically is depicted as follows in Figure 3.

**Figure 3**

*Sequential Composition of f and g*

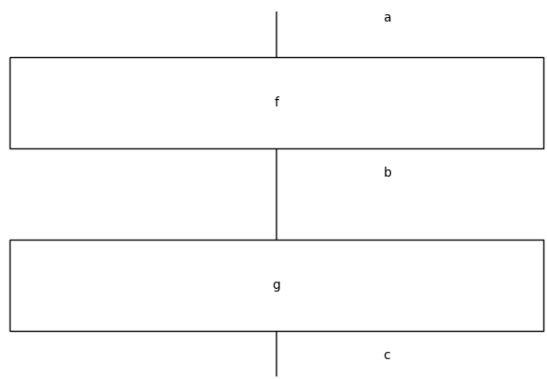

This is only the second stage of development of a diagrammatic depiction of category theory, and it already is easier to understand than using symbols. In this example, all one needs to do is remember that these morphemes are a funnel. Since time travels from top to bottom, object A enters morpheme *f* and outputs as B. Object B is then input into morpheme *g* and outputs as C. The logic flows much the same as when dropping a coin into a machine or dropping a marble into a Rube Goldberg machine. On the other hand, symbolic representation requires not only understanding of objects and morphemes, but also an understanding of what the colon (:), the arrow (→), and the newly introduced circle (∘) mean. Each of these characters introduce new readings that must be memorized for the symbolic representation to make sense. Recall the mnemonic used earlier for ∘ to be read as "after". The diagrammatic approach introduced here avoids any extra overhead.

Before continuing, let us demonstrate associativity for sequential composition using the diagram for the example above.[6] The symbolic equation for associativity is

---

[6] A more detailed proof can be found in Coecke's 2016 work.

$h \circ (g \circ f) = (h \circ g) \circ f$. Note that because time flows downward, there is no need to use parentheses. There is only one simplified graph required. See Figure 4.

**Figure 4**

*Sequential Composition Associativity Diagrammatically Depicted*

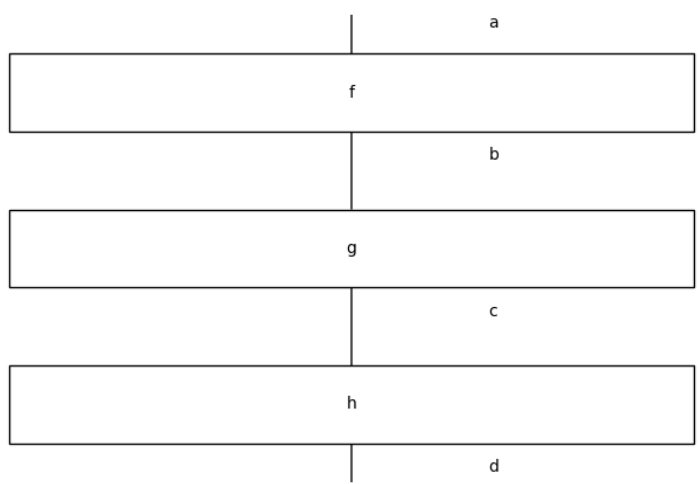

What about parallel composition? As it turns out, this is also very easily represented diagrammatically, as in Figure 5. Recall that the symbolic representation is $A \otimes B$, and the figure means "A while B" (Coecke & Kissinger, 2017). The diagram manages to capture the same idea without forcing one to understand the tensor product, or even the simplified understanding of the tensor product used in category theory—A's space added to B's space. The diagram shows both morphemes and all inputs and outputs plainly. Though the morphemes do not interact with one another, they do exist.

This picture represents the totality of possibilities as they relate to these morphemes

with only two boxes and four wires. As an aside, it is worth noting that associativity for

parallel composition can be demonstrated simply by graphing *g* on the left side of *f*

instead of the right side.

**Figure 5**

*Parallel Composition of f and g Diagrammatically Depicted*

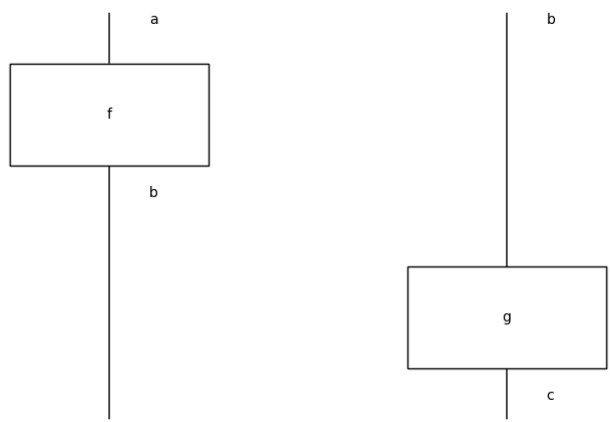

        To create graphs that represent quantum mechanical and linguistic ideas, the

final required diagrammatic component for string diagrams connects inputs to inputs and outputs to outputs. This is done using "*cups*" and "*caps*" (Coecke & Kissinger, 2017). They get their names from their appearance. Figure 6 depicts a cup, and Figure 7 depicts a cap. The X represents the original object, which is also referred to as the base type. The X.R and the X.L represent the right and left adjoint of the underlying matrix structure respectively. Beneath all these neat wires and boxes lies a rich world of linear algebra. When considering grammar, the underlying formulation is some mathematical grammar, such as Lambek's pregroup grammar or Combinatory Categorical Grammar—CCG (Coecke, 2020; Lambek, 1997, 2008; Yeung & Kartsaklis, 2021). This choice is typically not critical for later processing (Bolt et al., 2016).

**Figure 6**

*A Cup*

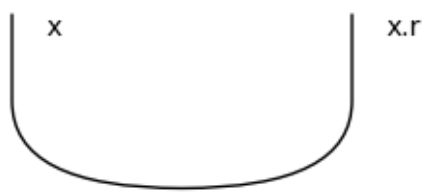

**Figure 7**

*A Cap*

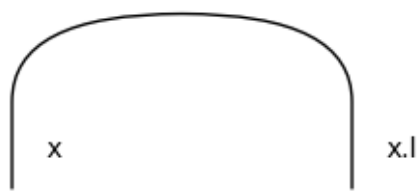

Now, to understand how cup and cap connections work, it is instructive to see an example. Figure 8 depicts cups and caps in the context of a larger diagram. Note that this toy diagram is based on the most often used example by Coecke et al. (2020). While it is true that this sentence provides a toy example, it, nevertheless, is a complete linguistic diagram of a simple *subject-verb-object* (SVO) sentence in English. Therefore, this diagram deserves some initial analysis and introductory explanation.

**Figure 8**

*Cups and Caps in the Context of an English Sentence*

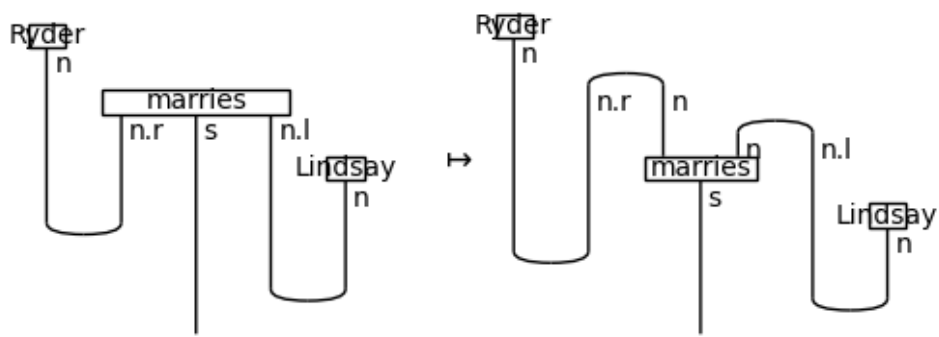

First, note that the left side and right side of the arrow are equivalent in meaning even though they are not equivalent in representation. The left side is simpler and requires only cups to complete the diagram. Each noun in the sentence is represented as a box—like a morpheme, also called a process[7], in Figure 1—at the semantic level, while the noun is assigned the pregroup grammar base type *n*. Transitive verbs also appear as boxes, but the wiring is more complicated. The transitive verb requires two nouns to be connected to it, one on the left and one on the right. Due to the rigid word order of English, the noun on the right side is the object of the verb, while the subject is the noun on the left side. The verb, then, must contain the right adjoint and left adjoint of the subject noun and object noun base types, respectively, and it must output some base type to indicate that the sentence is grammatically correct. This output type is referred to as *s* in the diagram. It is worth noting that all the wires are outputs—not inputs—because they are located on the bottom of the boxes. Again, thinking of the underlying structure of these diagrams as matrices is helpful here. The subject noun matrix and its right adjoint cancel out.[8] The object noun matrix and its left adjoint matrix cancel out. All that remains, then, is the sentence type. This is understood as proof that the sentence is grammatical. It follows that if these simplifications are not possible, then the sentence is not grammatically correct.

---

[7] Morphemes and their objects are often referred to as processes when the objects are envisioned as inputs and outputs (Coecke & Kissinger, 2017). This is simply because processes have inputs and outputs. It is a higher-level abstraction of the mechanisms at play.

      Next, the right side of the diagram follows all the same rules and simplifies down to a grammatically correct sentence as well. The difference is that there are also caps present in this diagram. Here it is easy to see the magical power of cups and caps. The difference on the right is that there is an additional noun type required to be an input into the transitive verb. Using the rule of string diagrams that allow outputs of one box to plug into outputs of another box, provides a simpler way to work with the

---

[8] This is the case in quantum computation because the matrices are necessarily unitary. The adjoint, then, is equivalent to the inverse.

underlying mathematics. This, ultimately, is the reason for cups and caps. Including them provides several elegant simplifications commonly referred to as the "yanking equations"[9] (Coecke & Kissinger, 2017). By leveraging the yanking equations, many diagrammatic simplifications become straightforward. To conserve space, diagrams in this work are displayed after applying the yanking equations going forward.

## Background: The Meta of the English Language

Some might consider English the lingua franca of the world as it is spoken by billions of people and recognized as the de facto language of international business and politics throughout the world. English is a non-agglutinative, West Germanic, Indo-European language that has absorbed components of many different languages—French, Latin, and Greek to name a few—into itself (Potter et al., 2020). Like most languages, it has evolved greatly over many centuries so that its contemporary form is very distinguished from its earlier forms. Even native speakers of the language struggle to discern texts written in its older forms.

---

[9] The yanking equations are also known as the "snake equations" (Al-Mehairi et al., 2017; Coecke et al., 2018b).

The grammar of English deserves some comment before proceeding. There are essentially three kinds of sentences: simple, compound, and complex. Simple sentences follow a very strict SVO pattern (Potter et al., 2020). The subject of the sentence is the actor, or the thing being discussed if the sentences is passive. The verb is the action and the dynamic core of the sentence. Without the verb, the subject and the object, also often called the complement, hang in a spacey void—a kind of abyss devoid of semantic meaning. In other words, without the verb, nothing ever happens. There are only players, but no play. The object is the thing that receives the action or is being acted upon, and it often has some supporting grammatical structures that are referred to as the predicate. A simple sentence must strictly follow the SVO word order or semantically the sentence will fail. An example of a simple sentence is as follows: "Mary birthed Jesus"[10] (*King James Bible,* 1769/2017, Luke 2:7). The grammatically prescribed word order in English is very rigid with only a few exceptions—fluidity in placement of adverbs and prepositional phases being two examples (Potter et al., 2020). Additionally, there are compound sentences that feature a direct object and either an indirect object or some predicate noun, verb, or adjective. This means there are multiple complements in a sentence with a single subject and one core verb. An example of such a sentence is as follows: "Jesus gave Mary to John"[11] (*King James Bible,* 1769/2017, John 19:26-27). Finally, there are complex sentences that contain multiple whole clauses, which follow the rules of the simple or compound sentences above mentioned. The simplest example here is "Mary birthed Jesus, and then Jesus gave Mary to John." It is critical to reemphasize that the structure of the English language is indeed very rigid. There is little that can be changed in terms of word order that does not also require some other sort of grammatical change to compensate.

---

[10] For clarity, this is not a direct quote of the text. The quotation indicates only the content of the example sentence.

## Literature Review: Brief Overview of QNLP Diagrams in English

It is worth the time to review relevant QNLP literature before introducing the Japanese language and its diagrammatic representations. Specifically, this section introduces work from studies performed on the English language. The goal here is to create a point of comparison with the Japanese diagrams to be introduced later. Particularly, it is crucial now to review word order and grammar.

---

[11] For clarity, this is not a direct quote of the text. The quotation indicates only the content of the example sentence.

In English, meaning is fundamentally tied to word order. Commonly, it is said that English follows a SVO word order. This is true, especially when referring to the kernel of an English sentence. Larger sentences require a more refined understanding of what precisely makes a subject and what precisely makes an object—often called the predicate. Much of the QNLP work has primarily focused on addressing the kernel of meaning. In fact, the examples provided in the Background: Category Theory section of this work do as well. See Figure 8 for an example generated with DisCoPy. A common example, with the same topological structure, from much of the literature is represented in Figure 9 (Coecke, 2016, 2020; Coecke et al., 2020, 2021). As mentioned before, the word order is lifted directly from English and into the diagram. This works to create a string diagram to calculate a semantic meaning precisely because the semantic structure of the sentence is built atop an underlying algebraic grammar (Bolt et al., 2016).

**Figure 9**

*English DisCoCat Diagram's SVO Word Order Core Topology*

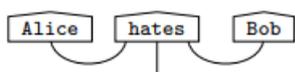

*Note.* This figure is featured in many of Coecke's works (2016, 2020; Coecke et al., 2020, 2021). Whenever words are included in a diagram, that diagram is functioning at

the semantic level. The algebraic grammar of choice is the foundation on which such a diagram is constructed.

The choice of grammar is flexible and an area of active research. Yeung and Kartsaklis recently based their work on a Combinatory Categorical Grammar (CCG)—Figure 10—where it is possible to curry the parts of speech defined by the grammar and arrange them into planar graphs[12] to produce grammatically correct sentences (2021). In Figure 10, "his hat" is connected to "put" via an underwire because it is the object of the verb. "Bruce" and the remaining unpaired underwires of "put" are curried over to "on", which accepts them as inputs—recall that quantum string diagrams accept inputs into their outputs as well to capture ideas like entanglement—and outputs a grammatically correct sentence.

**Figure 10**

*A Combinatory Categorical Grammar Implementation of DisCoCat*

---

[12] It is worth mentioning that the word order can be altered, and the meaning can be preserved. If this happens, then the graph is no longer planar, and the sentence would read "Bruce put on his hat". This does not change the SVO kernel of the sentence, however.

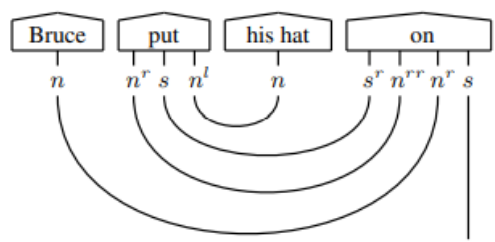

*Note.* This image originally comes from Lorenz et al. (2021). The most helpful part of this image for this work is that it clearly depicts both the semantic and the grammatical level of the graph. The semantic is depicted within the boxes, and the grammatical is depicted on the wires.

Most examples in the literature use Lambek's pregroup grammar, which Lambek produced as a replacement for earlier mathematical linguistic calculi (Lambek, 1997, 2008). See Figure 11, which is pulled from an experiment that tested these theories on real hardware (Lorenz et al, 2021). This sentence is grammatically identical to Figure 9. "John" is the subject, and "Mary" is the object. John "likes" Mary, meaning that he has some sort of fondness for her. The verb type is made up of a right adjoint and a left adjoint of the noun base type as well as a sentence base type. When the adjoints are joined with the base types, they contract, leaving only the sentence base type remaining. This confirms that the sentence is grammatically correct and, therefore, produces a semantically valid meaning. Pregroup grammars are often used because they are genius in their simplicity, easy to understand, and the foundation for many other category theoretical approaches that followed—the aforesaid CCG being a prime example.

**Figure 11**

*Example of Lambek's Pregroup Grammar with Cups and Caps as Underwires*

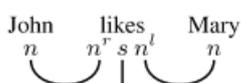

*Note.* This image originally comes from Lorenz et al. (2021). In this image, the boxes

are missing. They can be assumed to be present for simplicity.

      Now, DisCoCirc takes a similar approach but is intended to work with larger

bodies of texts. See Figure 12 for an example of an SVO sentence in DisCoCirc

(Coecke, 2020). Figure 12 contains two nouns: Alice and Bob. In a longer text, the

meanings of words, particularly nouns, are altered as the context begins to add nuance

to them (Coecke, 2020). In this case, Alice's meaning is updated to include the fact that

Alice is a dog. In kind, Bob's meaning is updated, defining that Bob is a person. The word order is preserved in the diagram because it is fundamental to English. Therefore, it is clear here that Alice is the one who bites Bob. Subjects appear to the left of verbs, and objects appear to the right.

**Figure 12**

*An SVO Sentence in DisCoCirc*

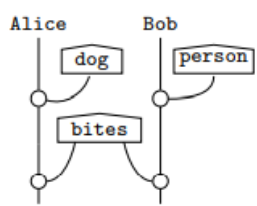

*Note*. This image originally comes from Coecke's 2020 work.

Now, to be clear, it is not the nouns' specific positions that determine left and right. It is the noun's associated wires. Pay special attention to the wire orders in Figure 13, which represents the same sentence as Figure 12. The nouns, their wires, and their modifiers have been completely swapped; however, the meaning is the same because the wires connect to the verb in the same respective spots. When taking these same concepts and applying them to languages with different grammatical structures, the underlying algebraic grammar must be modified. Indeed, many languages function entirely differently than English. When applying these concepts to such a language, distinct differences in the topologies of these diagrams will emerge, which is, in part, what the present study aims to illuminate.

**Figure 13**

*Figure 12 with Alice and Bob Swapped*

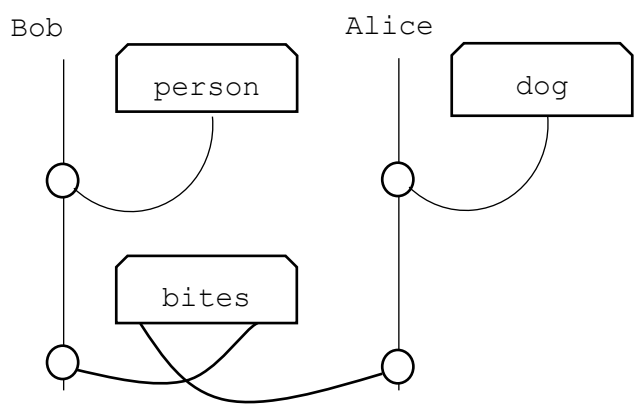

**Background: The Meta of the Japanese Language**

Japanese is one of the world's major languages with approximately 128 million speakers worldwide. These speakers primarily reside in Japan, but the popularity of Japanese entertainment and media has spread interest in the language abroad. The genetic origin of the Japanese language is a bit of a mystery (Shibatani et al., 2015). There is still active debate in scholarship regarding the origin of the language and whether there are any other languages closely related to it—aside from Ryukyuan.[13]

Like all languages, it evolved throughout time so that earlier forms of the language are difficult for contemporary speakers to understand without some degree of familiarization (Shibatani et al., 2015). Close interaction with China has repeatedly influenced Japanese over the course of many centuries. This has led to the adoption of kanji into the written language and the use of many Chinese loanwords. In more recent times, the influence of the West has led to an increase of English and other European loanwords being adopted into Japanese and spelled with katakana[14] characters.

The word order of Japanese, when compared with English, is remarkably fluid. General rules exist, and most often say Japanese follows a Subject-Object-Verb (SOV) word order (Shibatani et al., 2015). This is not entirely true (Dolly, 2019a). More importantly, it is not required to be true because the grammatical functions of words are marked explicitly in Japanese with special little words called particles. There are particles for marking the subject, the object, direction, and much more. Particles that appear at the end of sentences are used to perform functions like adding emotion, adding emphasis, soliciting a response, or turning the sentence explicitly into a question. Despite the presence of particles, it is incorrect to assume that word order does not matter at all (Dolly, 2019a). It is mandatory that a sentence be ended by a verb, adjective, or the copula, and it is generally accepted that the topic of discussion or the subject of the sentence resides near the beginning of the sentence or paragraph.

---

[13] Ryukyuan is a common way to refer to Okinawan, which is a dialect disparate enough from the rest of Japanese that it is often described as a separate language.

[14] One of the more incredible aspects of the Japanese language is that one is required to learn four different writing systems to be able to read fluently. These writing systems include kanji, hiragana, katakana, and romaji. First, kanji characters are symbolic, meaning that each character has a meaning unto itself. Kanji are the closest written characters to an ancient system such as Egyptian hieroglyphics still in use today. Second, hiragana characters are used to write native Japanese words and to add inflectable endings onto kanji characters. Third, katakana characters are used for loanwords, primarily from the West, and to write something emphatically. The latter use is like how italics are often used in English writing. Finally, romaji—the English alphabet—is used directly in many cases for convenience or for cool factor. The Japanese people tend to have a high view of the English language and, therefore, a desire to sprinkle romaji characters in their texts whenever appropriate.

## Japanese Compared to QNLP Literature

This section compares Japanese to English regarding the current state of QNLP research. The purpose of this section is to contrast Japanese with English concisely. Now, the concept of SVO sentence structure does not exist in Japanese. Recall that many believe that Japanese is often called an SOV language. In truth, two basic rules for word order exist (Dolly, 2019a). The first of these, as alluded to before, is that the last word in a sentence is the verb, an adjective, or the copula—the "engine" of the sentence (Dolly, 2019a). The second is words can only modify words that come after them. Typically, word modification is limited to words that are directly adjacent to one another, but adverbs and time-related words have greater flexibility. In practice, this typically results in an SOV ordering where the subject is mentioned prior to the object. Now, the key to Japanese sentence fluidity is that particles paired with nouns determine those nouns their grammatical role in the sentence. Depending on the use particles, Japanese sentences take different forms. The first of these can be thought of as self-move, while the second is other-move (Dolly, 2018b). Conceptually, self-move and other-move sentences differ in whether the action of the sentence is directed at the subject in the former case or the object in the latter case. This is roughly, though not precisely, equivalent to transitive and intransitive sentences in English.

      Self-move sentences are typically very simple. The complexity of the sentences, of course, varies; however, the kernel of the sentence is a subject, the particle が [ga], and a verb. Table 1 shows an example sentence.

**Table 1**

*Parts of Speech of a Self-move Sentence with English Translations*

| Subject | Particle | Verb | Possible Translations |
|---------|----------|------|------------------------|
| 猫 | が | 渡る | The cat crosses over. |
| Neko | ga | wataru | The cat will cross over. |

The sentence kernel is simple enough and easy to demonstrate. The subject "neko" is followed by "ga". The main role of "ga" is to mark the subject of a sentence, and that role is key. Many Japanese teachers incorrectly profess that "ga" can mark the subject or the object of the sentence, and that is not the case. This construction, then, grammatically indicates that the cat is moving itself across something—context would determine the identity of said something but perhaps a street or threshold—as opposed to being carried across said something by another subject. It is also worth noting that self-movement is not the same thing as passive voice. The cat in this sentence is acting upon itself. It is not being acted upon by another any other agent.

Other-move sentences provide the grammatical construction required to express actions done by subjects to objects. Therefore, in other-move sentences the subject is acting on something besides itself. Table 2 provides the counter example to that provided in Table 1.

**Table 2**

*Parts of Speech of an Other-move Sentence with English Translations*

| Subject | Particle | Object | Particle | Verb | Possible Translation |
|---------|----------|--------|----------|------|----------------------|
| 私 | が | 猫 | を | 渡す | I hand the cat over. |
| Watashi | ga | neko | wo | watasu | I will hand the cat over. |

In this sentence, the subject is 私 [watashi], which is one of the first person pronouns of Japanese.[15] Thus, "I" performs the action on the object, which is a cat in this case. The receiver of the cat is not specified in this sentence, which it is unimportant for this example, but the receiver would be dictated by the context since the receiver is often obvious. Suffice it to say that the entity being moved by the action is an

---

[15] Because Japanese has a multi-layered, complex system of honorifics, there are multiple first-person pronouns. This is a dramatic difference with the English language and warrants its own study on how to best incorporate worldly context to provide more fluent translations.

important concept built into Japanese sentences via verbs and supported by particles like を[wo].[16]

Once again, the rigid SVO structure of English is drastically different than the more fluid word order of Japanese. This means that any algebraic grammar, such as Lambek's pregroups, must be curated to support the Japanese language. Further, the self-move and other-move concept is critical to understanding a Japanese utterance, so a fluent translation must account for verb choice and the corresponding use of particles to correctly discern and translate meaning. The closest English grammatical constructions are transitive and intransitive sentences. It is worth investigating whether quantum translation programs could leverage density matrices, or some other method, to fluently translate these sentences by leveraging grammar and context. This work hopes to lay the groundwork for such an investigation in the future.

**Towards a Diagrammatic Taxonomy of the Japanese Language**

---

[16] The particle を is commonly pronounced like the English word "oh" despite the fact that it is spelled with the character for the "wo" sound. Essentially, the "w" sound is omitted, making the character silent in most utterances.

A taxonomy is a rigorous classification of items within a given category. In this section, diagrammatic depictions of self-move and other-move sentences are introduced. These diagrams will be created using both the DisCoCat and DisCoCirc methodologies, derived from category theory, for each individual grammatical construction. The intent of this section is to demonstrate the power of diagrammatic reasoning while introducing a small collection of diagrams for others to use in their own research. In time, this small collection can be built into a true taxonomy. This is important because once the topologies of the diagrams are identified, any sentence where words of the same parts of speech can be process regardless of the specific words chosen. The underlying grammar for the diagrams in this section is based on Lambek's pregroup grammar. Before proceeding, this grammar must be defined explicitly.

**A Pregroup Grammar for Japanese**

Pregroup grammars are logical grammars that were introduced and explained using Figure 8. As a brief review, the basic idea is that types are used to calculate whether a group of words, typically a sentence, is grammatical or not. In practice, left and right adjoints are grouped together with the base—non-adjoint—type. When this happens, the grouping effectively cancels out and the grouping contracts to a value of

one—effectively cancelling the value out of the calculation[17] (Cardinal, 2007). In the case of sentences, the goal is to contract all the types save for one—the sentence type—that ensures that the sentence in question is a grammatically correct utterance. What form might a pregroup grammar take for Japanese? Fortunately, some research into this question has been done. Cardinal defines the basic types as shown in Table 3 (2002).

**Table 3**

*Cardinal's Pregroup Grammar for the Japanese Language*

| Symbol | Part of Speech | Example | English Meaning |
| --- | --- | --- | --- |
| $\pi$ | Pronoun | 私 [watashi] | I |
| $\bar{n}$ | Proper name | 鈴木 [Suzuki] | The name "Suzuki" |
| $n$ | Noun | 猫 [neko] | Cat |
| $a$ | Adjective | 赤い [akai] | red |
| $a_n$ | Adjectival Noun | 好き [suki] | liked |

---

[17] A contraction refers to computations on simple types, represented here by a generic base type $t$, of the following forms: $t^l t \to 1$ and $t t^r \to 1$. Expansion is precisely the reverse: $1 \to t^l t$ and $1 \to t t^r$. These are the two fundamental operations performed on pregroup grammars (Cardinal, 2007).

| Symbol | Part of Speech | Example | English Meaning |
| --- | --- | --- | --- |
| $a_v$ | Adjectival Verb | 難しい [muzukashii] | difficult |
| $\alpha$ | Adverb | 新しく [atarashiku] | newly |
| $s$ | Statement without a germane tense [a] | ほら [hora] | Look! |
| $\bar{s}$ | Topic sentence | 私は私だ。 [Watashi ha watashi da.] | I am me. As for me, I am me. |
| $s_1$ | Non-perfective tense statement [b] | 彼が走る。 [Kare ga hashiru.] | He will run. |
| $s_2$ | Perfective tense statement | 彼が走った。 [Kare ga hashitta.] | He ran. |
| $q$ | Question | 彼が走るか。 [Kare ga hashiru ka?] | Will he run? |

| Symbol | Part of Speech | Example | English Meaning |
| --- | --- | --- | --- |
| $c_1$ | Nominative complement | が [ga] | Marks the subject of the sentence. |
| $c_2$ | Genitive complement | の [no] | Shows ownership |
| $c_3$ | Dative complement | に [ni] | Marks indirect objects |
| $c_4$ | Accusative complement | を [wo] | Marks the object |
| $c_5$ | Locative or Instrumental complement [c] | で [de] | Specifies the boundaries, range, or extant of an action. This can refer to a location or a tool. |
| $c_6$ | Ablative complement | から [kara] | Specifies movement away from a noun |

*Note.* This table is adapted from Cardinal's 2002 master's thesis. The collection of all these types is sometimes referred to as the "alphabet" of the grammar (Bolt et al., 2016).

These are also often referred to as "generators" (Duneau, 2021).

[a] Some utterances are not necessarily grammatically complete or lack verbs, but the meaning of the utterance is carried by context. Examples include single words, idiomatic expressions, and meanings heavily implied by use of particles without the explicit use of a verb. It is possible that CCG approaches to grammars that allow for planar graphs could also be applied in Japanese to address these utterances (see Figure 10). This is an avenue for further research.

[b] The term "perfective" here is used in widely the same way as past tense is used in English, and for the purposes of this work that understanding is sufficient.

[c] で, "de" is a particularly versatile particle. An in-depth discussion on how "de" should be handled is outside of the scope of the present study. For now, suffice it to say that the locative and instrumental complement, or case, will be conflated. An argument can be provided that such a conflation is perfectly valid in the Japanese language (Dolly, 2019b).

The types presented in Table 3 seem much more complicated than one might expect given that Japanese words reduce primarily to nouns, adjectives, and verbs (Dolly, 2019b). Indeed, reducing these types to a minimal subset would be a worthy endeavor for further research, that topic is beyond the present scope. Further, while terms like nominative, genitive, dative, accusative, and locative refer to Western grammatical ideas, they help when considering how to parse a Japanese sentence and translate it into the English language. Since these concepts aid in parsing, which is critical to the endeavor of machine translation, they are worth considering for now. It is worth noting that combinations of the base types from Table 3 can be used to represent adjectives and verbs in a variety of ways. The concept that allows for various types of words to be reduced is defined by Cardinal's initial partial order postulates found in Figure 14 (2002).

**Figure 14**

*Initial Partial Order Postulates of Cardinal's Japanese Pregroup Grammar*

$$s_i \to s \to q;$$
$$\bar{s} \to s;$$
$$n_v \to n \to \bar{n} \to \pi;$$
$$a_v \to a;$$
$$a_v \to s_i;$$
$$a_n \to a.$$

*Note.* This image originally came from Cardinal's master's thesis (2002).

These postulates show the partial ordering of one base type to another by means of the binary relation (Cardinal, 2002).[18] This comes into play when selecting pregroup grammar types for use in implementing a part of speech for a conversational grammar. A type on the right, such as *s*, can link with any type to the left of it, such as $s_1$, and those types will also contract. It is akin to an inheritance relationship between classes or an interface in a strongly typed, object-oriented programming language. The interface, whether explicitly added to the class or inherited from a parent, broadens the possible uses and application of the class. An illustrative example is in order. Let us create an imaginary device to demonstrate. For laughs, let us call it a Walton Device. In essence, a Walton Device is any modular item onto which any arbitrary number of arbitrarily shaped plugs can be grafted. One is depicted in Figure 15.

**Figure 15**

*A Walton Device with Undefined Types*

---

[18] This construction somewhat contrasts with the actual linguistic reality. Pronouns are simply nouns. Nouns are not pronouns. So, the construction of the pregroup grammar is somewhat out of alignment with the actual linguistic reality of Japanese. While it is functional, it would be worthwhile to pursue bringing these two views of grammar into better alignment.

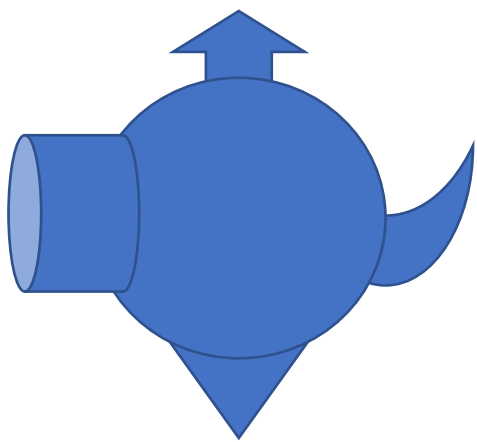

At first glance, the Walton Device depicted in Figure 15 looks a bit like an intergalactic polyhedral space pig. However, a more useful observation is that the main spherical body has precisely four uniquely shaped plugs protruding from it. It stands to reason that each of these distinct plugs has a matching, distinct receptacle counterpart into which it snugly fits. The Walton Device, then, is simply a graphical representation of the concepts of interfaces, types, and—in hardware—plugs. Each plug is a basic type, and all plugs on the same Walton Device are related according to some defined partial order, like those defined in Figure 14. Figure 16 shows the systematic construction of a Walton Device to represent a portion of Cardinal's noun rule from Figure 14.

**Figure 16**

*Development of a Walton Device for Noun Types Across Four Stages*

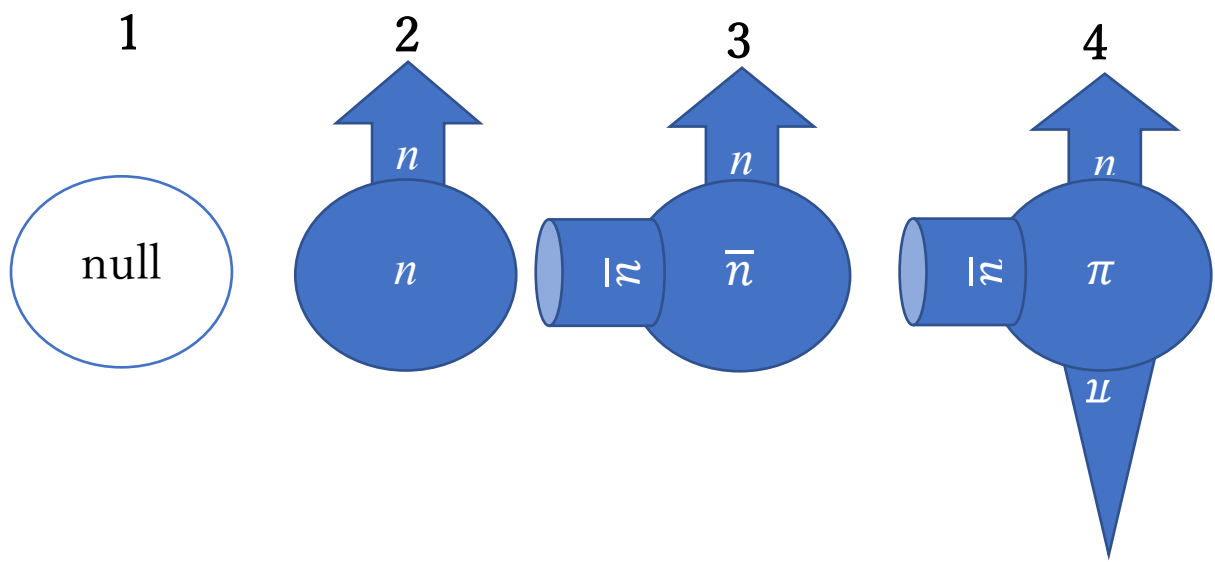

Starting from the left, Figure 16 develops a Walton Device from null to a representation of a subset the noun types in Figure 14. All Walton Devices start from null, which is represented at stage 1 with a circle because a circle has no pluggable shapes grafted onto it. For each stage, the most inclusive postulate type on the Walton Device is written in the circle as well. The most inclusive type is defined as that represented on the plug that the previous Walton Device from the same postulate is lacking. This represents the highest order type the Walton device can support if there is a hierarchy among the types postulated. If no such hierarchy exists, then the body of the device can be left blank or marked with an appropriate fundamental type—such as the empty set. At stage 2, then the noun interface is added to the top of the device. It is now able to plug into a noun receptacle, which in this case means the device can fulfill the role of a noun in Cardinal's grammar. Stage 3 adds the proper noun interface, thus increasing its potential connectivity, which is equivalent to saying its level of abstraction is increased. Finally, stage 4 adds the pronoun base type, which now gives this device three interfaces, maximizing the potential pluggability of this Walton Device. Therefore, in any place in the grammar where a noun-shaped receptacle would be it can be mated with any non-null Walton Device found in the hierarchy of Figure 16. This is the power of interfaces, the power of types, and the power of abstraction all wrapped up into one visual idea. It will soon be demonstrated that this degree of abstraction is extremely useful when working with Japanese particles.

Let us diagram some sentences using Cardinal's grammar to demonstrate this idea. See Figure 17, which handles the case of self-move sentences and juxtaposes each sentence with the corresponding Walton Device for the noun.

**Figure 17**

*Juxtaposed Self-Move Sentences with Walton Devices to Compare Noun Types*

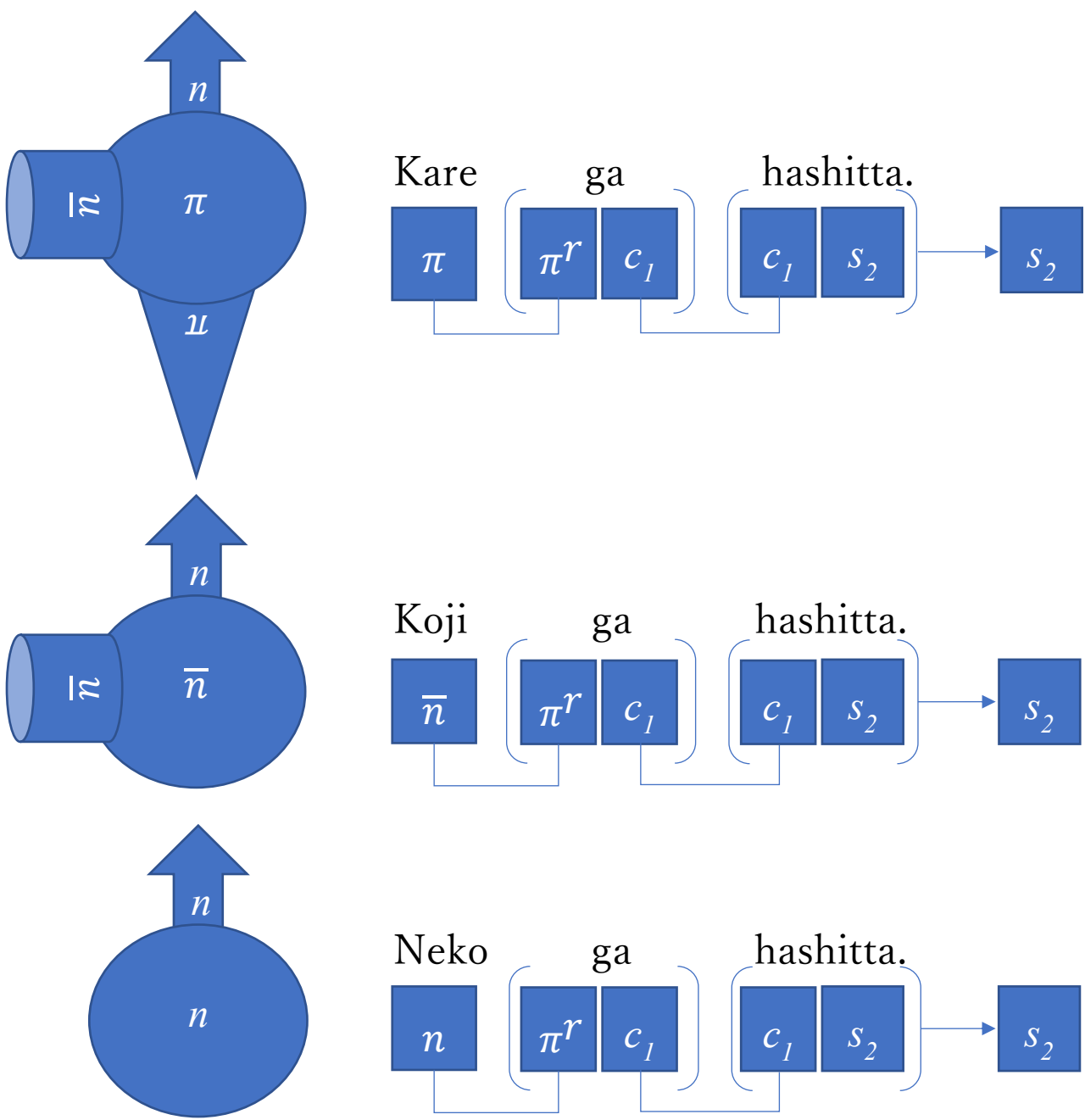

First, consider the top sentence of Figure 17. This sentence uses the third-person male pronoun as the subject. Since a pronoun is used, the $\pi$ type is assigned to it. Comparing the top sentence with the middle and bottom sentences illustrates the partial ordering and why Walton Devices are helpful visual aids. The subject of the second sentence is a proper noun, while the final sentence is a normal noun. All three of these noun base types are the subject of their sentences because they are paired with the "ga" particle. In accordance with Cardinal's partial ordering via the binary relation, the Walton Devices for the nouns have broader interfaces as one reads from the bottom of Figure 17 to the top.[19] Therefore, "neko", "Koji", and "kare" can all be paired with the constructed type for "ga" and the right adjoint *simple type* in the "ga" construction simplifies to 1, which cleanly removes it from the calculation. This leaves the nominative complement simple type $c_1$ unpaired. Complement types such as this are paired with the verb in the sentence, thus providing a link through the particle between the noun and the action of the verb. After this link is satisfied, the perfective tense sentence type $s_2$ remains, which satisfies the grammatical rules of both the logical grammar and Japanese semantics. Therefore, this sentence is well-formed and legal.

---

[19] The metaphor implies that the plug needs to mate with a receptacle. Depicting a receptacle is much more difficult with an image, so for simplicity, base types are written on the plugs even though the plugs technically should all have a superscript "r" after them. The right adjoints would be plugs in this metaphor.

Before concluding this section on pregroup grammar types, consider an other-move sentence. Everything previously illustrated with Walton Devices still applies, so Walton Devices have been omitted here to conserve space. Other-move sentences have an object, which introduces another particle into the grammar. It is instructive to see how this new particle is constructed using pregroup types.

**Figure 18**

*Type Grammar Representation of an Other-move Sentence*

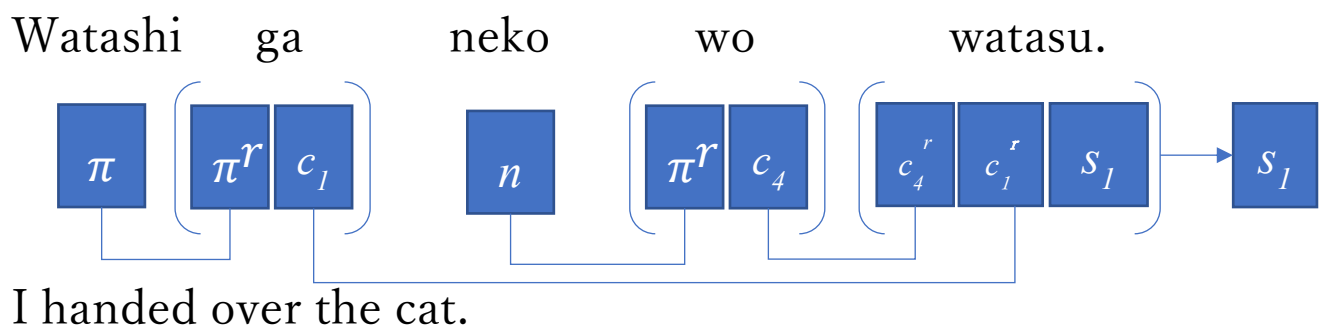

Figure 18 provides a second example of the usage of "ga", and its typing remains consistent. This suggests, correctly, that a particle has one simple type definition per one grammatical usage. This does mean that there is potential for other type constructions of these particles that would produce legal sentences (Cardinal, 2002). There are also constructions that are used more often than others. For the present study, this single construction of "ga" is sufficient. Another observation to make is that the right adjoint is very commonly used in Japanese. This is due to the heuristic that a given word modifies the word that immediately follows it in the grammar, which is not shocking since Japanese is an agglutinative language. Thus, most word pairings will be a word with its right-adjoint since the right-adjoint follows its paired word. Now, "wo" is used to mark the object of the verb. Here, there is a clear subject, and a clear object. Their interaction is, of course, defined by the verb. This necessitates[20] the inclusion of another simple type in the verb type construction for this sentence to cancel out the accusative complement type introduced by "wo". The verb is in the non-perfective tense $s_1$, so a non-perfective, legal sentence is what remains after the calculation. Using this grammar as the foundation, it is a simple matter to extend these ideas into diagrammatic representations using DisCoCat and DisCoCirc, which also extends beyond the realm of grammar and into that of semantics. Let us proceed to that end.

**Enter DisCo Diagrammatic Representations**

---

[20] See Cardinal's 2002 work for a slightly different approach that adds an additional sentence simple type to "ha", the topic marker, to create a legal sentence without adding a third type to the verb. This approach allows the result to boil down to a topic sentence $\bar{s}$.

From this point forward, quantum diagrams are introduced. Everything that precedes this section provides adequate background for understanding the underlying rationale behind these diagrams from a linguistic perspective. As certain key points become relevant, they will be re-emphasized to help facilitate understanding. Additionally, any general observations indicate some larger principles and heuristics will be highlighted and named as they occur.

**Moving the Self**

This section discusses how self-move sentences with predicate adjectives work, sets the stage for a direct comparison between a self-move and other-move sentence later, and consolidates all the diagram types for a self-move sentence into one section.

*Verb*

Before proceeding, revisit Figure 17. This figure depicts Walton Devices for self-move sentences with a noun, proper noun, and pronoun, in juxtaposition, all of which feature the verb 走った [hashitta]. With all the groundwork already prepared, it would be criminal to not build the DisCoCat and DisCoCirc diagrammatic representations of one of the sentences depicted here as an introduction. First, the DisCoCat diagram is depicted in Figure 19.

**Figure 19**

*DisCoCat Representation of a Basic Self-move Sentence with a Verb*

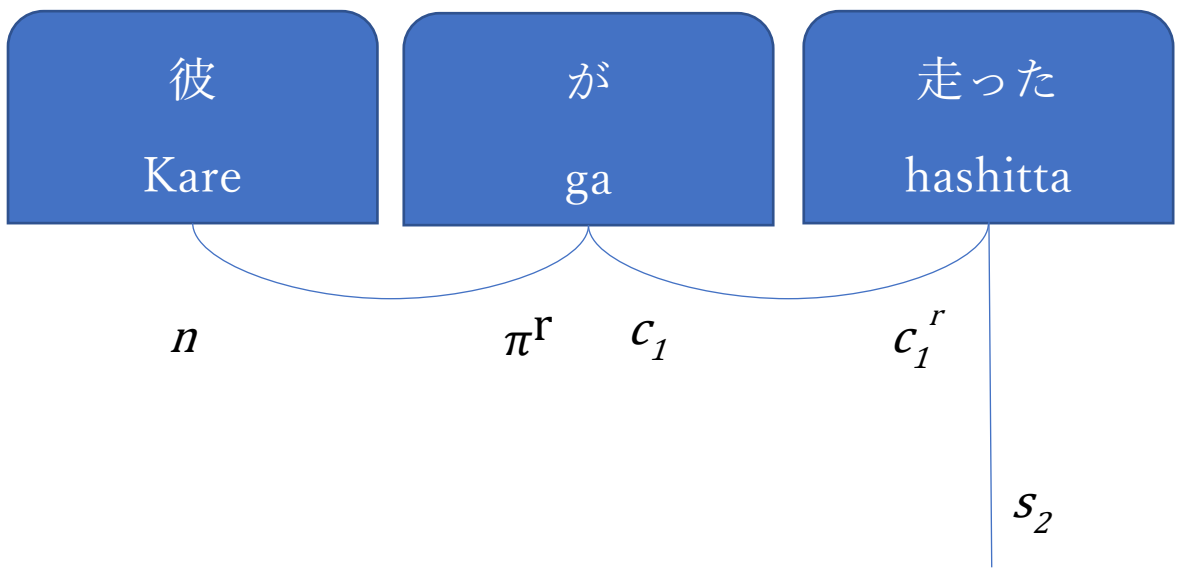

Figure 19 depicts a string diagram that is sufficient for implementing simple quantum mechanics and algebraic linguistics. This is the first depiction of such in the Japanese language, and this is one of the simplest diagrams that could be created. Even still, it warrants comments. The pregroup grammar provides the mathematical substructure to ensure that the connections are grammatically legal, as has been discussed earlier. The boxes represent the semantic level of understanding beyond the grammar substructure and is implemented via a quantum state (Coecke et al., 2021).

Each cup satisfies and symbolizes a matrix and its adjoint being multiplied together and resulting in a 1 value. This is realized by quantum measurements, or Bell tests (Coecke et al., 2021). It is, therefore, negated and provides no additional effect on the calculation. The final remaining type is some variety of the *s* or the *q* base types. These demonstrate that the sentence is grammatically legal and, therefore, makes sense semantically. Now, this explanation should sound familiar. It is, in essence, the same explanation offered when discussing underwires in pregroup grammars. To be clear, the major difference is the inclusion of the quantum states and measurements that require an additional level of implementation beyond that of grammatical linking (Coecke et al., 2021). Without quantum mechanics, these boxes would simply be boxes. It is astounding that with such a small visual step, linguistics gets quantum.

Now, here are some general observations about Japanese DisCoCat diagrams that derive from Figure 19. First, because Japanese is a very noun heavy language, there are going to be a lot of $\pi^r$ simple types encoded into particles. Why is this? Because a particle always succeeds the noun it suffixes, is directly adjacent to said noun, and particles abound in proper Japanese sentences. Further, using the $\pi^r$ type ensures that the particle can satisfy a standard noun, proper noun, and a pronoun. There is no reason to adjust the encoding due to the type implication imposed by the partial ordering previously illustrated with Walton Devices (see Figure 17). This observation is known as the *Pi are Numerous Principle*.

Second, the persisting sentence type that remains after the pregroup calculation tends to reside in the verb, adjective, or copula that ends the sentence. The final word in a Japanese sentence is the driver of meaning. All other words in the sentence are somehow related to the action or statement of being made by this final word. This observation is known as the *Train Engine Principle* since the engine sits at the end of the train and pulls it along the track[21] (Dolly, 2018a). It is worth noting, however, that there is an exception to this principle, one of which is particularly notable and will be introduced later. Now, it is important to notice that the final word's simple type definition is flexible. Therefore, while it can be said that a verb or a predicate adjective may have a typical simple type definition based on its normal—most frequent—usage, it cannot be said that the type definition will always be the same since the use of adverbs and additional particles are typically accounted for within the verb or adjective being modified. In addition to this observation, it is also true that most Japanese DisCoCat diagrams will be planar without the need for currying (see Figure 10). This observation is further supported by the next.

---

[21] Using train terminology to name the principles is an ode to Cure Dolly, who passed away this year. Much of her life was spent trying to teach Japanese to native English speakers in a more organic way, so naming things after her lessons is done here as a simple way to honor her memory.

Third and lastly, since most of the time Japanese nouns tend to modify the words that immediately follow them and that particles attach to the words that immediately precede them, there will be many cups connecting two adjacent quantum states in the diagram. As such, this observation is known as the *Train Car Principle*. Ultimately, this means there are minimal opportunities for graphs to become non-planar when diagramming out sentences in Japanese.[22] Before proceeding, it is worth discussing how particles seem to buck the trend of words modifying the word that proceed them? Is a particle not modifying the word that precedes it? This is a matter of perspective. It may be more helpful to consider the noun as modifying the particle that follows it. This produces an effect that allows for the particles to be explicitly translated into English using compound nouns or relative clauses. For example, a literalistic translation of Figure 19 that implements this approach would yield a translation like the following: "He, that is the subject, ran." If the relative clause makes the point too murky, then consider this translation "He-subject ran." Thus, what kind of subject is it? It is the "he" kind. So, in this way, the noun is making explicit some quality of the subject of the sentence. Thus, the words on the left side of the sentence continue to inform those on the right side of the sentence even if they are particles. If it is helpful, consider employing this perspective to make the meaning flow and connections between words more uniform. If not, think of particles as suffixes instead of independent words. From this perspective, suffixes will attach to words from the right side, while whole words will continue to modify the words that follow them—proceeding from left to right. This latter approach is the more native of the two.

---

[22] The notable exception could be a large leap in a topical sentence from adverbs that modify the final word of the sentence.

What about DisCoCirc? Figure 20 introduces a very simple DisCoCirc representation to match Figure 19. Because there is only one simple sentence, this diagram is essentially analogous to the "Hello World" program of DisCoCirc. Indeed, this example could not be simpler and remain functional. "Kare" is the noun, which is represented by the wire. The wire flows through "hashitta", a quantum state, which updates the meaning of "kare" to mean the "he who ran". In this way, a transitive verb updates the meaning of the noun just as an adjective does. Since the verb here is a quantum state and not an effect, the noun is not terminated and continues to persist, awaiting its next meaning update as the text continues.

**Figure 20**

*A "Hello World" Example of DisCoCirc Diagrams*

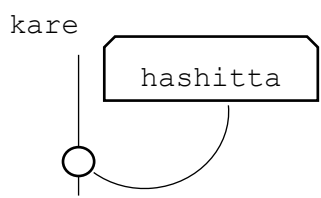

*Predicate Adjective*

Now, let us consider predicate adjectives in self-move sentences. One of the defining traits of Japanese is the topic marking particle は [ha], which is very powerful and multifaceted. It has the specific function of singling out the topic, but it also has the broader function of denoting difference, contrast, or non-inclusion.[23] The ability of the Japanese speaker to explicitly define a topic allows the speaker to change the *locus of*

---

[23] The sister particle of は [ha] is も [mo], which denotes inclusivity. "Mo" is often translated into English as "too" or "also".

*speech*, meaning that the perspective of the situation—where the speech is situated—can be shifted from one person, place, thing, or action to another with great ease without changing the grammatical subject of the sentence containing the topic. The topic, then, functions at a higher level—above the level of a single sentence—than the grammatical subject does. How does this work? Consider the sentence introduced in Table 4. While this sentence only contains seven words, it is sophisticated enough to demonstrate a crucial point.

**Table 4**

*Self-move Sentence's Parts of Speech Featuring a Predicate Adjective*

| Noun | Particle | Noun | Noun | Particle | Adjective | Copula | Translation |
| --- | --- | --- | --- | --- | --- | --- | --- |
| 私 | は | オレンジ | 色 | が | 好き | だ | I like orange. |
| Watashi | ha | orenji | iro | ga | suki | da | |

      First, notice that there are two particles in this sentence. From left to right, the first is the topic marker, and the second is the subject marker. Obviously, then, the topic and the subject of the sentence are two different entities. In this case, the topic of the

sentence is the speaker, and orange-color is the subject of this specific sentence. Again, topics are special because their reach extends beyond the realm of a single sentence and into larger blocks of text. The extent of a topic's reach is often referred to as a comment (Dvorak & Walton, 2014). In this case, the reach of the topic is the same as that of the subject since they are both confined to a single sentence example. So, then, how are the topic and the subject different? Are there two actors in this sentence? Plainly, no. The topic defines the perspective of the sentence in a case such as Table 4. The comment that follows after the は [ha] is a sentence unto itself, and it is framed within the view of the topic. Perhaps comparing progressively more literal translations will make this phenomenon more apparent. Table 5 juxtaposes three potential translations for easy comparison.

**Table 5**

*A Few English Translations of Table 4 Juxtaposed*

| Simple | More Detailed | Literal |
| --- | --- | --- |
| I like orange. | I like the color orange. | As for me, the color orange is likeable. |

The literal translation is a far cry from the simple translation. The latter tends to be the one taught to beginner-level students of Japanese. Notice how the adjectival noun 好き [suki], meaning "likeable", is translated as a verb.[24] Notice also how the subject of the sentence is conflated with the topic in the English translation. In fact, the simple English sentence is slightly more ambiguous. Is the "orange" here a color or a fruit? A native English speaker would be able to decipher this based on the singular "orange" instead of the plural "oranges", but perhaps a non-native reader would struggle to tell. The more detailed translation resolves the ambiguity, but there is still grammatical confusion regarding the subject, topic, and the translation of an adjectival noun into a verb. Therefore, the literal translation is provided to untie all these conceptual knots: "As for me, the color orange is likeable." Now, this sentence is not natural in English, but it does demonstrate, via the use of the comma and the preposition "for", the framing of the sentence from a certain perspective before proceeding to the body of the sentence. There is no ambiguity now. The subject is orange-color, and it is described by the predicate adjective "likeable". This, then, is what is meant by shifting the locus of speech with ease.

---

[24] This is called an adjectival noun here for expediency, but since it is a so-called "na-adjective", it is really just a noun that requires a special version of the copula, called な [na], to modify words that succeed it in the sentence (Dolly, 2017).

Having demonstrated that predicate adjectives are employed to reflect some quality about a subject from the perspective of the topic, let us now consider the algebraic grammar of such a sentence. This representation is found in Figure 21.

**Figure 21**

*A Self-move Sentence Logical Grammar Diagram with a Predicate Adjective*

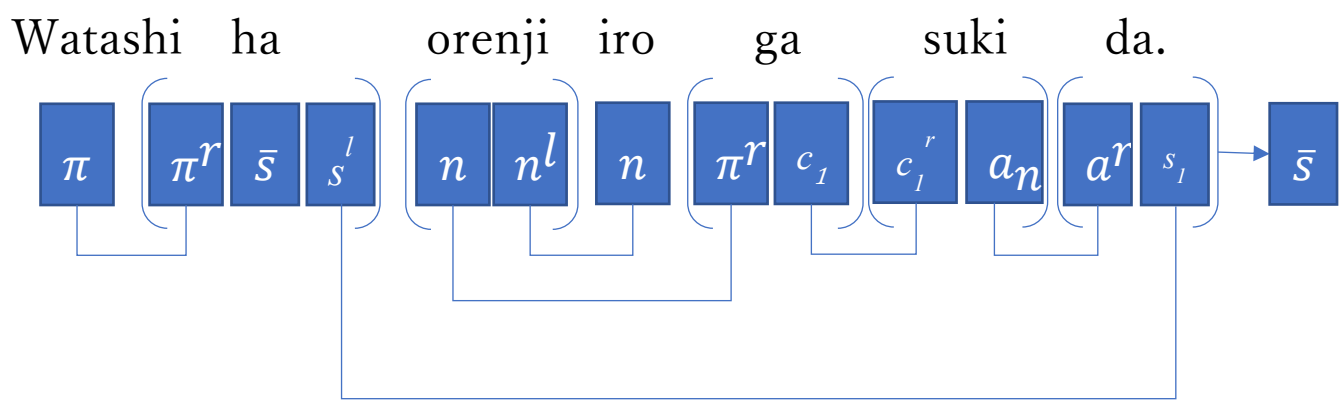

The analysis pertaining to Table 5 above is made evident in Figure 21. The "ha" makes the sentence a topical sentence. It also divides the sentence into two pieces like an equal sign does in equations. There is everything that comes before the "ha", which is the locus of the speech—the perspective from which the rest of the sentences comes or the situation to which it refers. Then, there is the sentence proper containing the subject as marked by "ga". Notice that this example uses the adjective types from Table 3 and the adjectival version of the copula. This same sentence's semantic meaning would be represented in a DisCoCat diagram as shown in Figure 22.

**Figure 22**

*DisCoCat Representation of a Self-move Sentence with a Predicate Adjective*

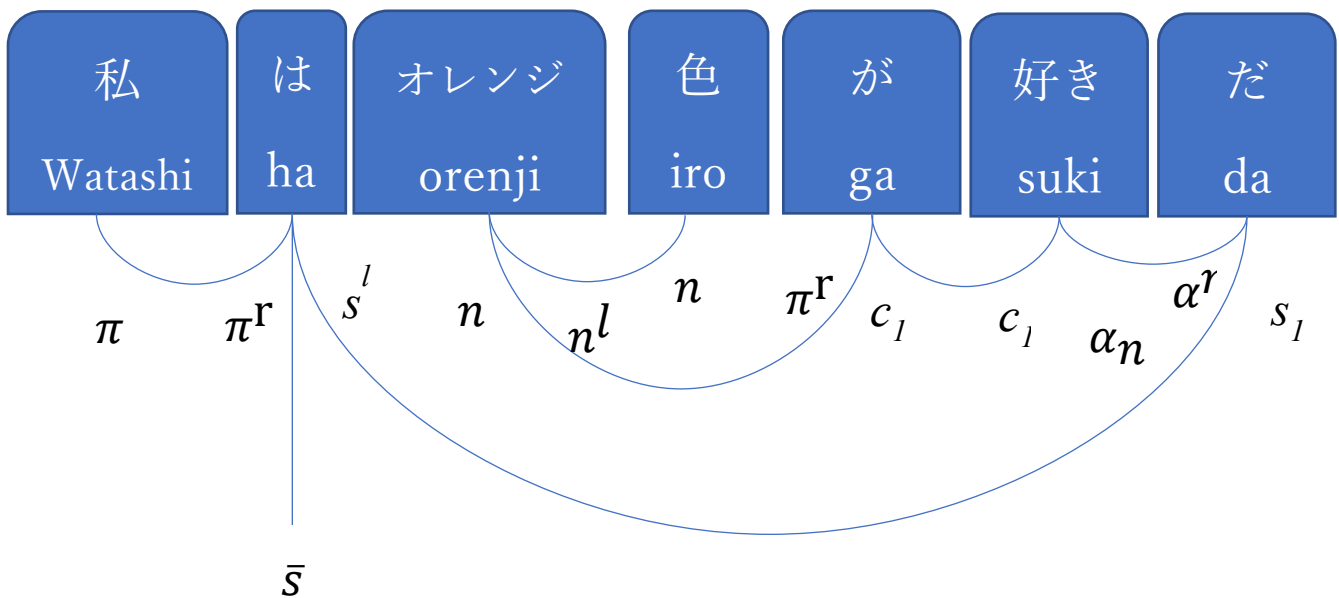

The DisCoCat diagram again quantizes the grammar to yield semantics. However, Figure 22 is differs from Figure 19. One obvious difference is that this diagram contains an adjective; therefore, the diagram contains adjective types. Specifically, it contains the adjectival noun base type and the right adjoint as a simple type. Adjectives in Japanese generally will be one of three types, and these three types are named based on their function in a sentence. It will be either a noun modifier, a noun, or a verb. The first of these cases is the noun modifier, which is the archetypal and essential case of an adjective. It modifies some person, place, or thing by attributing some quality, limit, count, or the like to it. In the second case, the adjective is simply functioning as a noun. In this case, the adjective has no morphological forms or transformations. Instead, a な [na] is affixed to the end of the word to modify the word that follows (Dolly, 2017). Lastly, the verb case refers to sentences where an i-adjective appears in the predicate position. These adjectives can be morphologically transformed to indicate positive, negative, non-perfective, and perfective tenses. Thus, they are behaving as verbs and might as well be considered a limited class of verb. Figure 22 represents the adjectival noun case, which has a unique partial order that reduces the adjectival noun to a noun: $a_n \rightarrow n$ (see Table 3; Figure 14). After this reduction occurs, the adjectival noun can be treated like a noun in the grammar in all other cases.

The first difference leads logically to the second. Instead of a verb, this sentence contains the copula だ [da]. The copula represents all possible versions of "is", "am", and "are" from the English language in one easy to use package. Thus, it is employed to create a grammatical statement of equivalence. Additionally, the polite and honorific forms of the copula can be used to raise the politeness level of a sentence. In the case of Figure 22, it is functioning only as a statement of equivalence and nothing more. The copula is an interesting case because in extremely casual Japanese, it is possible that the copula might be dropped so that the sentence is easier to say. In this case, it is treated like a particle. The reason why omitting the copula still carries the meaning is because of its ubiquitous use. Like が [ga], the subject marking particle だ [da] is present even when it is not spoken by the sheer power of implication. Because Japanese is a high-context language, it is simply assumed to be there even if unuttered. Thus, it is likely that parsing programs would benefit from inserting the missing copula at the end of the sentence when it is omitted, which would provide a degree of uniformity across sentences. This assertion remains yet untested and is an avenue for further research.

Now, what about the Train Engine Principle? Figure 22 seems to violate the heuristic that the sentence type is carried by the final word in the sentence. As mentioned earlier, there is an exception to the Train Engine Principle, but it is only an exception in part. In reality, it is still true that the sentence in Figure 22 carries the sentence type in its final word; however, Figure 22 also contains a topic marking particle は [ha]. That topic marker is the kicker. Recall from the discussion on Figure 21 that the topic marker divides the sentence into two chunks, identifying the left side of the sentence as the topic of the following comment. The right side of the sentence is its own self-contained sentence even if the subject is implicit and omitted. It turns out, when a topic marker is used, the topic sentence base type is what remains after the types resolve. Now, just as it makes perfect sense that the Train Engine Principle asserts that the sentence type resides in the last word of the sentence because it bears the meaning, it is equally true that the topic marker should bear the meaning of a topical sentence. This is because a topical sentence defines a perspective that informs all the sentences that follow until the topic is changed again. It is, therefore, fitting that the topic marker override the Train Engine Principle, since the latter functions at only sentence level. This observation is known as the *Non-Logical Primacy Principle* because the topic marker is considered a non-logical particle, while the subject marker and others are logical particles (Dolly, 2018c). Again, logical particles function at the sentence level while non-logical particles have a larger purview and provide different shades of meaning from their logical counterparts.

One final observation on Figure 22 is that it introduces left adjoint simple types. Unlike right adjoint simple types, which are included to resolve a previously introduced type that has no pairing, left adjoint types anticipate types to come. There are a couple of instances of this in Figure 22, but a single example should suffice. So, for instance, consider the cup connecting the topic marker to the copula. The topic marker is defined by three simple types: the right adjoint of the pronoun, the topic sentence base type, and the left adjoint of the sentence. The left adjoint cannot stand on its own, meaning that the topic marker[25] is anticipating the sentence to follow. This result is again fitting since the topic marker is forecasting a statement that provides information about the topic or from the topic's perspective.

---

[25] Because Japanese is such a high-context language, it is possible for the topic marker to end a sentence. The result is an implied question. For example, the sentence 私は [watashi ha]? Literally means, "As for me?" Because the rest of the sentence is omitted, the assumed portion of the sentence is assumed to be an inquiry. This assumed portion of the sentence is very strong and nearly always understood by the interlocutor. In this example, it is akin to saying, "What about me?" in the English language.

Next, consider a DisCoCirc diagram. Normally, the idea in DisCoCirc is to take the important nouns, persist them, and specify the relationships between them. These relationships are typically captured by verbs or adjectives, which then link the two nouns together or link directly to the noun respectively. Recall Figure 12 for an example in English. English, however, does not have a topic marker. There is, therefore, no direct parallel in the English literature on how to represent this. One idea is as depicted in Figure 23.

**Figure 23**

*Proposed DisCoCirc Diagram of a Self-Move Sentence with a Topic Marker*

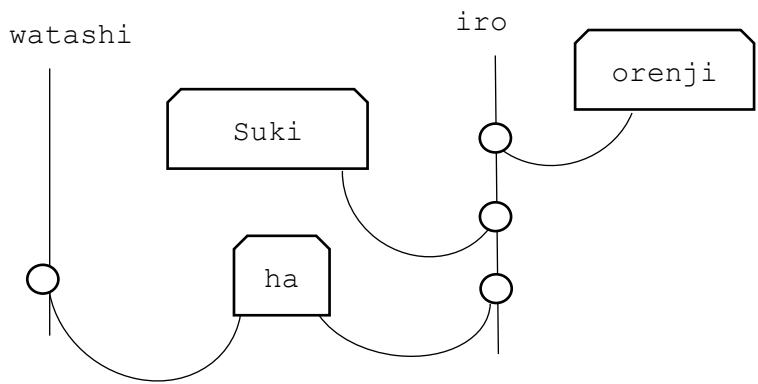

Let us consider what this Figure 23 is attempting to convey. The two selected nouns[26] of importance are "watashi" and "iro". The former means "I", and the latter means "color". This seems reasonable since the interlocutors are likely discussing their favorite colors. It makes sense that the term color would persist through the discourse and receive meaning updates. Here, the color is described as "orenji", meaning orange, and "suki", or "likeable" in English. Now, the first question is, how to link this meaning back to "watashi". The proposal in this image is to consider the topic marker akin to a transitive verb. Transitive verbs link a subject and an object together and update the meaning of both nouns accordingly. Figure 12 also includes an example of this happening in English. The question is, ultimately, whether this is an accurate representation of what is naturally occurring in Japanese speech. The entirety of the idea expressed in the sentence is clearly to update the meaning of "watashi". After all, each sentence until the topic change is grammatically required to be some comment related to the topic. An interesting problem here is how to ensure the sentences that follow continue to update the topic even though it is not present. One likely approach is to persist the current topic and add the implied topic back into each sentence when parsing. Next, recall that the sentence containing the topic marker is typically divided into two halves, with the latter half being both a comment on the former and its own independent sentence. For now, including the topic marker in this way is a potential solution that would involve a minimal amount of special consideration when working with the Japanese language. More tightly formulating the solution to this problem, working with larger blocks of text, and testing out potential solutions is an area for further research.

---

[26] For simplicity, the term "noun" is used to refer to nouns, proper nouns, and pronouns since the difference ultimately does not matter in Japanese grammar or DisCoCirc as either are formulated in this section.

### *Same Old Verb, Different Forms*

Obviously, verbs are another part of speech that can produce a self-move sentence. Figure 19 is one example where exactly this occurred. The difference in this section is that the idea of a single verb having a self-move and other-move form is introduced, discussed, and demonstrated. Such a verb one stem but slight morphological differences that signify whether the verb moves the self or moves the other—whether the verb is akin to an intransitive verb or a transitive verb in the English language. This section introduces a self-move form of one such verb and serves as a bridge from the self-move portion to the other-move portion of this work as well.

As always, the natural grammar of the sentence is introduced first. To begin, then, consider Table 6. There's nothing particularly unique or outstanding about this sentence, but it does accomplish two things. First, simply note that the verb is 決まった [kimatta], which is derived from 決まる [kimaru]. This verb means "decide" and is self-moving or intransitive. That means it is linked directly to the subject and not an object in the sentence. Second, Table 6 represents the first use of an adverb type. Its function is, naturally, to modify the use of the verb.

**Table 6**

*A Self-move Sentence with a Verb*

| Noun | Particle | Adverb | Verb | English Translation |
|------|----------|--------|------|---------------------|
| それ | は | もう | 決まった | That was already decided. |
| Sore | ha | mou | kimatta | |

Next, the algebraic grammar diagram is shown in Figure 24. Again, this is a very straightforward representation. It is worth mentioning a couple things, however. First, observe how the adverb type, represented by an alpha, modifies the verb that directly succeeds it. This is the standard case, and there is nothing new in this example.[27] Second, note that the use of "ha" is not the only way to render this sentence. Figure 24, as it stands, would be used to change the topic to "sore", meaning "that", as opposed to something else. If there is no need to change the topic, a "ga" would be used instead. This would change the representation of the verb to that of Figure 25 instead.

**Figure 24**

*A Pregroup Grammar Representation of a Self-Move Sentence with a Verb*

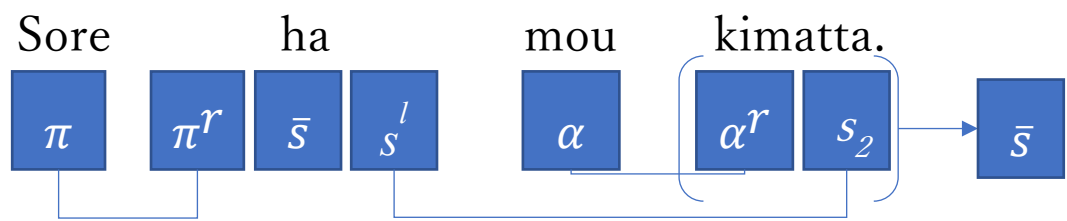

That was already decided.

---

[27] Adverbs are freer in terms of sentence structure than most other parts of speech in Japanese. The adverb is often the first word of the sentence. When this happens, it is typically represented with the left adjoint of the sentence type to indicate that it is modifying the whole sentence rather than just the verb. In effect, these two types of modification are not so different given the way Japanese grammar parses.

The Figure 25 case simply shows that self-move verbs can accompany が [ga], or は [ha]. The sentence will ultimately fulfill a different purpose depending on the employed particle.[28] In this case, the $c_I^r$ simple type must be added to the verb to account for the use of "ga". Therefore, this distinction is critical when considering fluent speech and translations. Forcing a translation algorithm to discern between topics and subjects is of paramount importance if a translator is going to be truly fluent.

**Figure 25**

*Alternative Self-move Verb Options with Subject Marker Instead of Topic Marker*

---

[28] When "ha" is used and there is no "ga" present, the implied subject is equivalent to the topic. In this case, the "ga" is dropped for convenience. Nevertheless, the implied subject's presence remains in the sentence.

With Adverb

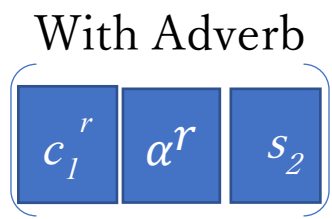

Without Adverb

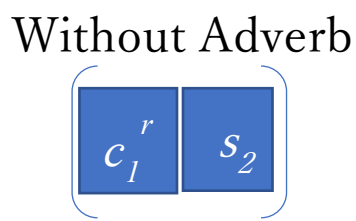

*Note.* This diagram also demonstrates the type encoding differences when an adverb is used. The adverb type is represented by the Greek letter alpha.

Now, returning to the representation found in Figure 24, let us proceed with the DisCoCat representation of this sentence. It is found in Figure 26. Again, this graph is included for comparison purposes. There is nothing in this graph that has not already been introduced, but Figure 26 is included for completeness and is useful for comparison..

**Figure 26**

*DisCoCat Representation of a Sentence with a Self-move Verb*

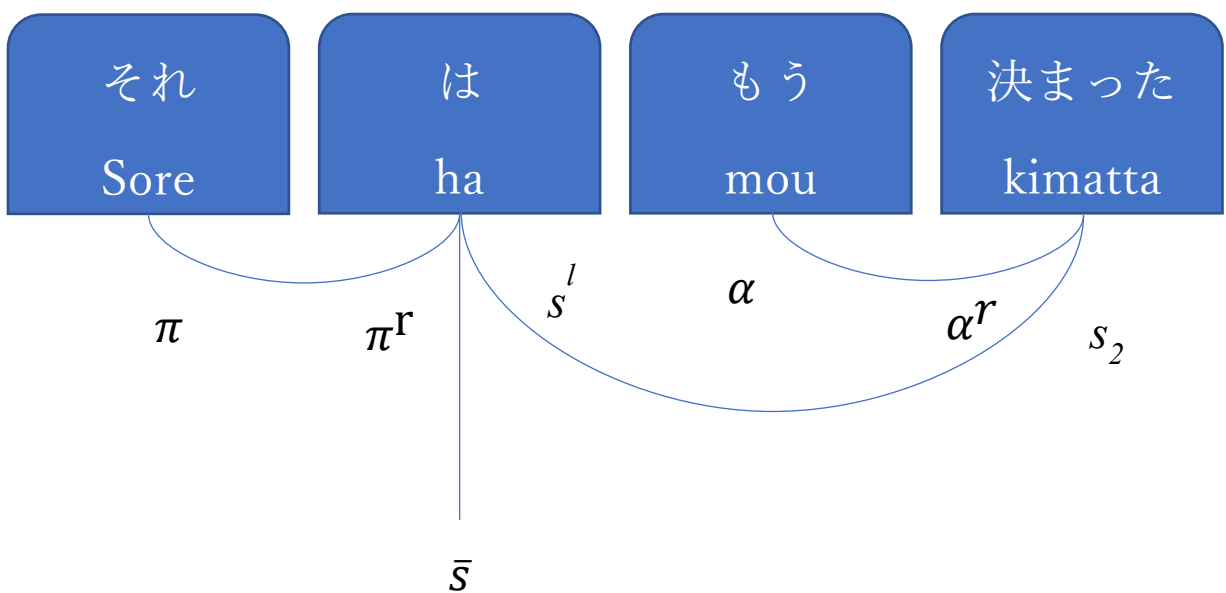

The DisCoCirc diagram for this sentence is included here for completeness, though this graph would be unlikely to be used. Because "sore" is a pronoun, the noun to which it refers would likely be the noun of importance that would be preserved and updated through the larger discourse. Nevertheless, Figure 27 documents the diagram that would suffice, if required. It is worth noting that this DisCoCirc diagram is structurally equivalent to Figure 20; however, including the adverb as part of the updating process can capture a more nuanced idea of decision.

**Figure 27**

*A DisCoCirc Diagram of a Sentence with a Self-move Verb*

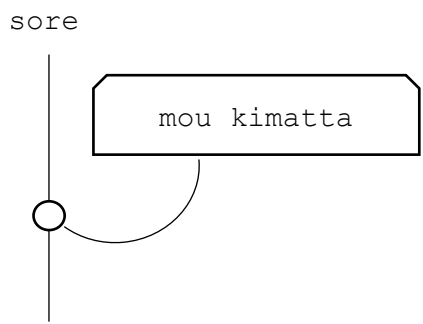

*Note.* The self-move verb functions equivalently to an adjective. See Figure 20 for a comparison.

**Moving the Other**

Other-move sentences are more complicated than self-move sentences by their very nature. Every sentence in Japanese requires at least an implicit subject marked by が [ga]. This subject and the subject marker itself are often dropped in casual speech, but even if it is not spoken, it is both grammatically and contextually implied. This is possible due to the high-context nature of the Japanese language. Especially deft speakers can say an incredible amount with a minimal number of words.[29] An other-move sentence not only includes the subject marker, but it also includes the object marker を [wo].[30] The sentence then, by definition, must include an additional noun, an additional particle, and all the grammatical connections required to support them. An example is provided here to analyze other-move sentence structure in more detail. Furthermore, this section will feature the other-move version of the previously discussed verb 決まる [kimaru], which is 決める [kimeru].

---

[29] Exploring this idea is itself an area for further research. For a quantum computer to provide consistent and fluent translation outputs, it may be necessary for the translation algorithm to insert implicit subjects and particles. It is worth studying specifically what this overhead might amount to when performing translations and how well inserting explicit grammar where it is intentionally omitted might scale to large bodies of text. Additionally, it is worth investigating what the minimal type alphabet and grammar rule set might be to handle extremely casual speech. Both avenues of research are beyond the scope of the current study.

Table 7 introduces the other-move sentence in natural Japanese grammar. As before, simple parts of speech are used to emphasize that most Japanese words fall into the broad, easy to identify categories. In addition to an other-move verb, Table 7 also includes a relative clause. Relative clauses in Japanese are too large a topic to discuss here in detail, but the important thing to remember is that Japanese creates relative clauses be modifying nouns on the right with verbs on the left.[31] In this way, an entire sentence worth of meaning can be directly attributed to a single noun.

---

[30] In fact, it is possible for there to be multiple object markers when the same subject is performing the same action on multiple objects.

[31] Many have discussed the wiring of relative pronouns in English, which requires special consideration (Coecke et al., 2018c; Sadrzadeh, 2013, 2014). Japanese, thankfully, does not require special consideration regarding this topic because relative clauses are not constructed with relative pronouns.

**Table 7**

*An Other-move Sentence with a Verb*

| Noun | Particle | Noun | Particle | Verb | Noun | Particle | Verb | English Translation |
|---|---|---|---|---|---|---|---|---|
| かれ | は | 薬 | を | 付ける | こと | を | 決めた | He decided to apply the medicine. |
| Kare | ha | kusuri | wo | tsukeru | koto | wo | kimeta | |

When translated literally into English, the obvious subject of Table 7 is "Kare, who decided". The content of that decision is made explicit by the relative clause "kusuri wo tsukeru koto". One way to translate this idea more literally into English is to refer to it as an event or an action. Transforming a verbal idea into an explicit noun is one of the primary roles of "koto" in Japanese. Table 8 is provided to help make the function of "koto", meaning "thing", "matter", "situation", or "event", easier to grasp. Based on Table 8, it is easy to see the function of "koto" in Table 7 even though the literal translation is very unnatural in English. "Koto" takes a clause with a verb and makes it a generic relative clause. It nominalizes a verb so that it can play the role of a noun in the sentence. Why is this important? Verbs are unable to accept logical particles[32] as suffixes. Therefore, there must be some way to make a verb into a noun to fulfill grammatical duties in conjunction with particles. This is done in English using "-ing" as a suffix. In this Japanese sentence, it is often done using "koto".[33]

**Table 8**

*Juxtaposed Potential Translations of Table 7's Other-move Sentence*

| Literal English Translation | Fluent English Translation |
| --- | --- |
| As for him, he decided the apply the medicine matter. | He decided to apply the medicine. |

---

[32] Verbs can, however, take sentence-ending particles as suffixes. Like は [ha] and も [mo], sentence-ending particles are non-logical in the sense that they do not define any specific grammatical relationship between words in a sentence. They, instead, perform functions like providing emotion and soliciting feedback.

[33] This can also be done with the particle の [no]. The two sometimes overlap enough in meaning that the choice does not particularly matter, but often one sounds more natural than the other depending on the context and content of the sentence. Covering this in more detail is beyond the scope of the current discussion, but it is a potential area for further research. Capturing the subtlety in semantics is critical.

Now, Table 9 provides a direct comparison between the self-move and the other-move versions of the sentence-ending verbs in Table 6 and Table 7. Comparing these two verbs directly is instructive.

**Table 9**

*Direct Comparison of Self-move and Other-move Forms of a Verb with Same Stem*

| Self-Move Form | Other-Move Form |
|---|---|
| 決める | 決まる |
| kimeru | kimaru |

There are a couple of things to notice in Table 9. First, note that the kanji character used in both words is the same. The roles of the kanji in the word is to convey the basic, or core, meaning of the word and to represent the word's stem. The kanji can be compounded by combining with a second kanji, or the kanji can be supplemented with Japanese kana characters. This latter case is what is done in true adjectives and in verbs. Thus, the attached kana characters are used to shape the meaning of the verb in some way. The second point of importance here is that the second character in both words begins with the "m" sound. Now, the Japanese syllabary consists of columns of five vowel sounds and the consonants that can be prefixed onto the vowel to make a legal utterance. In Japanese, there are many verbs built on the same kanji that differ only in that the middle syllable's vowel sound in one is an "e" sound and an "a" sound in the other.[34] This is a telltale sign that one is the self-move form of a verb, and the other is the other-move version.[35] While it is usually true that the "e" sound signals an other-move form and the "a" sound signals a self-move, this is not always the case. This is one of the relatively few places where Japanese is not entirely consistent (Dolly, 2018b). The use of particles is, once again, the ultimate arbiter on whether a verb is self-move or other-move. Why is this important? Primarily because understanding intricacies such as this in Japanese will help determine the optimal number of qubits to use in translation circuits. The morphology of Japanese ought to be studied to define the level of qubit granularity needed to optimize translation. More qubits provide more flexibility but also more overhead. The opposite is also true. The question, which is another area of potential research is, do translation algorithms provide more fluent outputs when operating at the word level or at the morpheme level? The answer remains to be seen.

---

[34] A fascinating topic related to this one is the 五段 [godan] verb. "Godan" verbs have five different forms of the same core of a word. The core is indicated by a kanji character and, often, a hiragana character. The five different vowel sounds each represent a different basic meaning alteration applied to the core idea of the word. Quantum implementations of "godan" verbs is another potential avenue for further research beyond the scope of this work.

[35] This is not the only potential difference between self-move and other-move verbs, but it is the difference that applies in this example. Another common difference is whether the final kana character is a す[su] or a る[ru]. A classic example of this difference would be 返す[kaesu] and 返る[kaeru]. The former is typically other-move and the latter is self-move (Dolly, 2018b).

Next, let us move on to the algebraic grammar analysis of the sentence from Table 7 to see the underpinnings of the logical grammatical structure. Figure 28 was devised to do just that. This example sentence is once again topical. The topical construction was chosen to keep the self-move and the other-move sentences more similar.[36]

**Figure 28**

*Pregroup Grammar Diagram of an Other-move Sentence*

---

[36] The non-topical construction would result in an other-move verb that links to the が [ga] particle. This particle bears the nominative type, which is $c_1$. Because the verb is also an other-move verb, it also must link to an accusative case particle with type $c_4$. The verb in Figure 28 is in perfective tense, so the complete type definition for the verb in the non-topical construction would be $c_1^r c_4^r s_2$.

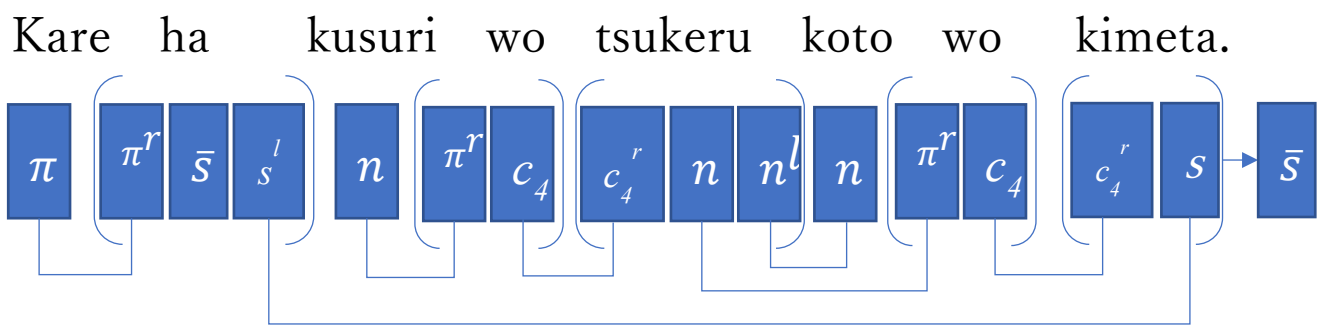

He decided to apply the medicine.

Because this sentence is topical also, it also breaks down into two halves like an equation. The right half is a fully grammatical sentence with an implied が [ga] that has been dropped since the subject and the topic are the same word, namely "kare". The relative clause in the sentence uses the noun こと [koto] to nominalize the clause, and the particle を [wo] is suffixed onto "koto" to indicate the accusative case. The verb "tsukeru" relies on a useful type definition strategy to connect anything to a succeeding noun. This strategy, known as the *Noun Sandwich Principle*, relies on the left adjoint's ability to forecast a type in the sentence to create a noun sandwich (see Figure 21). This sentence contains the accusative particle twice, which might seem alarming at first; however, since the accusative particles are in separate clauses, it causes no ambiguity and, therefore, no problem. The utterance remains perfectly comprehendible. It is the accusative particle that follows the "koto" that links to the other-move verb, namely 決めた [kimeta].

The next diagrammatic evolution is the DisCoCat diagram. Figure 29 shows the DisCoCat representation. By now, it is not surprising. It is a perfect extension of the pregroup grammar diagram found in Figure 28 into the realm of semantics.

**Figure 29**

*DisCoCat Diagram of an Other-move Sentence with a Relative Clause*

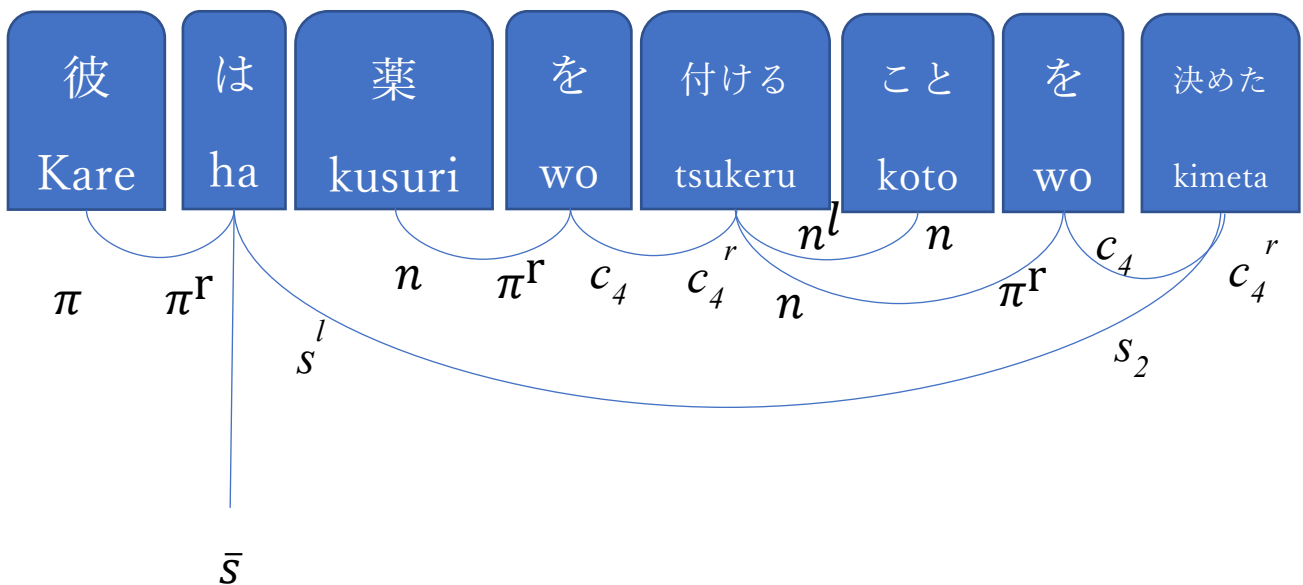

Figure 29 demonstrates several of the same principles that have been introduced before. The topic marker は [ha] again demonstrates Non-Logical Primacy. In this example, however, there is a large leap of five words, non-inclusive, to the main verb—the would-be train engine. Again, because the sentence is topical, the Train Engine Principle is superseded. The Train Car Principle is evident as there are many cups between adjacent quantum states, or words. The Noun Sandwich Principle is also evident. It is here applied to join the other-move verb 付ける [tsukeru] to the nominalizer こと [koto]. The noun sandwich has an intriguing quality in that it relies on a cup that skips over a noun to connect to the subsequent particle. This gives the effect of suffixing the particle onto the verb in Figure 29, which is precisely what "koto" intended to do. Suffice it to say that the DisCoCat representation is perfectly capable of representing self-move and other-move sentences. It is instructive to compare the diagrams with one another to pictorially show the underlying grammatical mechanisms at play and how these mechanisms inform the topology of the semantic diagrams. This is particularly true when comparing diagrams containing the self-move and other-move versions of the same base verb. Ultimately, the choice of verb form determines whether the train cars can be towed well; therefore, this type of consideration is of utmost importance when musing over the idea of fluency.

Lastly, a DisCoCirc diagram is provided in Figure 30. This diagram is representative, meaning it is one possible diagram that might exist. It is very likely that こと [koto], would not be considered a primary noun because it is used primarily out of grammatical necessity.[37]

**Figure 30**

*DisCoCirc Diagram of an Other-move Sentence*

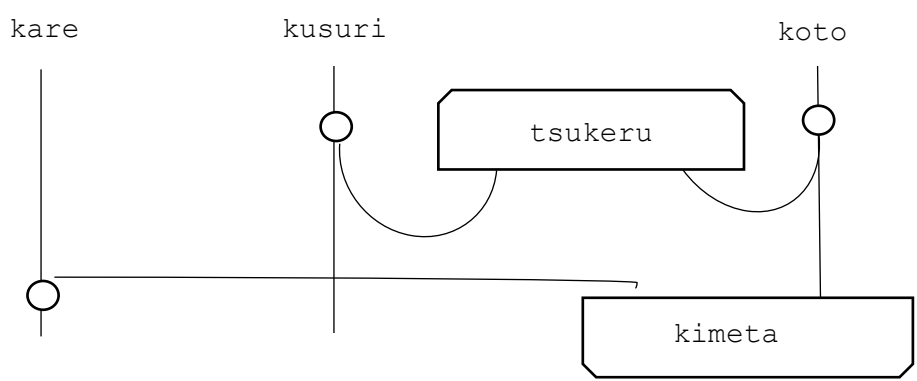

Nevertheless, for this example, "kare", "kusuri", and "koto" are chosen as the primary nouns, and each is granted its own wire. These nouns mean "he", "medicine", and "thing" respectively. Because of the realization that the "koto" wire may not be permanent, it ends by feeding directly into "kimeta", the other-move verb that links with "kare". Terminating "koto" in this way is instructive. This termination is depicted graphically by mirroring the quantum state across a horizontal cross section. This is the diagrammatical representation of an effect, which in DisCoCirc representation represents an imposition of a subject onto an object (Coecke, 2020). The noun representing the object has its wire terminated as part of the effect, which is a quantum measurement. Now, reasoning backwards—up—through the diagram, a rough translation of the bottom and left portions would be something like "he decided the thing". The quantum effect used to represent "kimeta" indicates that deciding something in this graph terminates the wire. The upper and right portions link "kusuri" through "tsukeru" to "koto". This, clearly, is the portion of the diagram that represents the relative clause. The "thing" being discussed here is the "application of medicine thing". Thus, when both sides are taken as a whole, the upper-right side is completed first. This is represented by the verb "tsukeru", meaning "apply to something" in this case., being listed higher than "kimeta", which means "decided". So, the meaning of "koto" is updated first, then the meaning of "kare" is updated last. "Kare" and "kusuri" continue to persist farther into the narrative—are outputted through the sentence as represented by a process. "Koto", on the other hand, does not. This termination of noun wires is an important distinction between self-move and other move sentences. The former applies attributes to the subject, while the latter acts on objects.

---

[37] Though, it is possible that nominalizers like こと [koto] and の [no] could be selected as primary nouns over a very large text since these nouns will appear over and again. They are exceptionally common. Whether there is any benefit to including them is an avenue of further research and experimentation.

## Conclusion

In a world where quantum computers are capable of error correction, assuming this requires the devices to also have a suitable number of qubits, translation devices using QNLP will be superior to classical machines with standard methods of NLP; therefore, the power of quantum computers may be the key to achieving unprecedented translation accuracy and even fluency. Achieving this goal is still beyond the horizon, but enough of the groundwork has been laid to consider how such a dream may be achievable. That groundwork has been the subject of this research and is summarized here.

First, category theory-based approaches to QNLP show much promise because they can represent complex mathematical concepts and calculations as simple pictures with varied boxes and wires. One no longer needs to be a quantum physicist to represent and solve problems that reduce to quantum mechanics. It is even possible to create these diagrams using software tool like DisCoPy, which is a fantastic boon. Furthermore, they are also a very natural extension of pregroup grammars and partial orders, which makes quantum calculations approachable to algebraic linguists. Quantum computing, then, is becoming increasingly democratized by these novel approaches.

The English language has been the primary research focus when working with category theory-based diagrams. This leaves the research open to biases when considering other languages. The English language perspective is one out of a great many of languages in the world. As a logical and low-context language with a strict word order core, the infamous SVO ordering, it is dramatically different from agglutinative languages that demonstrate another word order and have different grammatical rules for verbs. Studying the Japanese language then, precisely because it is so different from the English language, leads to remedies for some potential oversights that ought to be addressed in a QNLP pipeline. The present work addresses self-move and other-move verbs, the baseline grammatical structures that support them, and the string diagrams that represent their semantics. Additionally, this work serves as a primer to the behavior of particles in DisCoCat and a first speculation into that of DisCoCirc.

As far as visual representations are concerned pregroup grammar, DisCoCat, and DisCoCirc diagrams for Japanese self-move and other-move sentences were introduced and discussed. The study of DisCoCat diagrams yielded some observable principles that aid in making sense of the topologies of the diagrams. To provide the same for DisCoCirc would require investigations of much larger bodies of text. These DisCoCat principles are as follows:

1. The Pi are Numerous Principle: That, due to the partial order of nouns as depicted by Walton Devices and the frequency of use of particles in the Japanese language, diagrams of sentences with any degree of complexity will contain many pronoun types, which are represented by $\pi$. These are mainly manifested as the right-adjoints of $\pi$, shown as $\pi^r$, hence the wordplay in the choice of "are" in place of "is". The latter would be grammatically correct, but much less fun.

   b. The Train Engine Principle: That the final word of a Japanese sentence is always the mover of the sentence. The *s* base type is encoded into this final word. It is, therefore, like a train engine pulling a series of train cars. The exception to this rule is topical sentences, which have their own overriding principle.

   c. The Train Car Principle: That Japanese modification flows from left to right in single hops much like how train cars are linked one to another to form a proper train. The exceptions to this principle are topical particles and adverbs, which, despite having a standard position, have more word order freedom in Japanese. Nevertheless, this general observation is a foundational heuristic of these graphs.

d. Non-Logical Primacy Principle: That non-logical particles, such as the topic marker は [ha] will supersede the Train Engine Principle. The *s* base type is encoded in the non-logical particle in this case. The reason for this is the topic can influence meaning beyond the sentence that contains it. It affects a larger portion of the text, which is known as the comment (Dvorak & Walton 2014).

e. The Noun Sandwich Principle: That adjectival nouns, which rely な [na] to connect to the noun that follows, verbs in relative clauses, and other words that function as prefixes are encoded with $nn^l$ and that these types will be joined by a n base type from the right, the adjacent word being modified, by a cup. The result is that three noun types are used together across two words. This creates a sandwich of noun types.

This work also introduces many avenues for continued research into Japanese. One first order concern is the granularity with which diagrams are constructed. Japanese contains not only words, but morphemes. It is not at all obvious whether constructing diagrams at the word level or the morpheme level produce more accurate and more fluent translations. This becomes particularly important when working with causal and passive verbs as morphemes are critical to correctly parsing the meanings of these verbs. Concerning parsing generally, Japanese, as a high-context language, drops many of its crucial grammatical elements when spoken. This includes the subject marker が [ga], which is the linchpin of a sentence. Parsing methods will be crucial as a preprocessing step to producing DisCo diagrams. It is likely that implied elements will need to be made explicit during the parsing process. Obviously, there is the question of what typical diagrams might be created in the ZX-Calculus, or other potential diagrammatic representations, and whether there are any useful topological principles that may be gleaned from them. It follows that typical quantum circuit topologies remain to be demonstrated based on the baseline of this work. Additionally, other algebraic grammars remain to be tested. It is certainly possible to test alternative grammars for potential hidden benefits as well as minimize Cardinal's alphabet to a streamlined set. Of course, there are many other grammatical constructs in Japanese that remain to be studied. Adverbs, adjectives, and relative clauses are mentioned here, and this work would serve as a good foundation to launch into those investigations in more detail. There are also many grammatical choices with implicit meanings that warrant ferreting out. A great example here is the use of こと [koto] or の [no] for nominalizing verbs. Naturally, there is also the topic of logical negation and how to best represent that concept in Japanese. Negation in Japanese is handled by the adjective ない [nai], which functions quite differently than the "not" of English. There is also much more to be said about representing particles in DisCoCirc diagrams. Particles function very uniquely when compared to English grammar. In the same vein, Japanese also has layers of politeness that are heavily, heavily informed by worldy context. It, then, is absolutely crucial to investigate ways to incorporate worldly context into translations so that the proper level of politeness—in addition to conversational negation—is produced by the translation (Rodatz et al., 2021). The translation cannot be fluent without addressing this concern. Lastly, any one of these observations could be the subject of experimentation with publishable results. Many of the finer points of quantum translation will require quantitative and qualitative study to fine tune the algorithms and processes.